\documentclass{article}

\usepackage[preprint]{neurips_2024}

\usepackage{amsmath}
\usepackage{amssymb}
\usepackage{mathtools}
\usepackage{amsthm}
\usepackage{natbib}
\setcitestyle{numbers,square}
\usepackage{multirow}
\usepackage[ruled]{algorithm2e}
\usepackage{bbding} %for \CheckmarkBold

\usepackage[utf8]{inputenc} % allow utf-8 input
\usepackage[T1]{fontenc}    % use 8-bit T1 fonts
\usepackage{url}            % simple URL typesetting
\usepackage{booktabs}       % professional-quality tables
\usepackage{amsfonts}       % blackboard math symbols
\usepackage{nicefrac}       % compact symbols for 1/2, etc.
\usepackage{microtype}      % microtypography
\usepackage[dvipsnames,cmyk,table]{xcolor}         % colors
\usepackage{arydshln}
\usepackage{xspace}
\usepackage{float}
\usepackage{wrapfig}
\usepackage{lipsum}

\usepackage{color, soul}
\usepackage{colortbl}
\usepackage{setspace}

\usepackage{hyperref}       % hyperlinks

\hypersetup{
  colorlinks=true,
  citecolor=[rgb]{0,0.55,0.27},
  urlcolor=RoyalBlue,
  linkcolor=[rgb]{0,0.55,0.27}
}

\newcommand{\AgentGym}[0]{\textsc{AgentGym}\xspace}
\newcommand{\AgentTraj}[0]{\textsc{AgentTraj}\xspace}
\newcommand{\AgentEval}[0]{\textsc{AgentEval}\xspace}
\newcommand{\AgentEvol}[0]{\textsc{AgentEvol}\xspace}

\title{\AgentGym: Evolving Large Language Model-based Agents across Diverse Environments}

\author{
Zhiheng Xi\thanks{{ }{ }Equal Contribution.} \ \thanks{{ }{ }Correspondence to: \texttt{zhxi22@m.fudan.edu.cn, \{tgui, qz\}@fudan.edu.cn }}  \ , Yiwen Ding$^*$, Wenxiang Chen$^*$, \\
\textbf{Boyang Hong, Honglin Guo, Junzhe Wang, Dingwen Yang, Chenyang Liao,}\\ \\
\textbf{Xin Guo, Wei He, Songyang Gao, Lu Chen, Rui Zheng, Yicheng Zou,}\\ 
\textbf{ Tao Gui$^\dag$, Qi Zhang$^\dag$, Xipeng Qiu, Xuanjing Huang, Zuxuan Wu, Yu-Gang Jiang} \\ \\
\large \text{Fudan NLP Lab \& Fudan Vision and Learning Lab} 
}
\definecolor{lightgray}{RGB}{211, 211, 211}
\definecolor{lightfont}{gray}{0.3}

\begin{document}

\maketitle

\begin{abstract}
Building generalist agents that can handle diverse tasks and evolve themselves across different environments is a long-term goal in the AI community. Large language models (LLMs) are considered a promising foundation to build such agents due to their generalized capabilities. Current approaches either have LLM-based agents imitate expert-provided trajectories step-by-step, requiring human supervision, which is hard to scale and limits environmental exploration; or they let agents explore and learn in isolated environments, resulting in specialist agents with limited generalization. In this paper, we take the first step towards building generally-capable LLM-based agents with self-evolution ability. We identify a trinity of ingredients: 1) diverse environments for agent exploration and learning, 2) a trajectory set to equip agents with basic capabilities and prior knowledge, and 3) an effective and scalable evolution method. We propose \AgentGym, a new framework featuring a variety of environments and tasks for broad, real-time, uni-format, and concurrent agent exploration. \AgentGym also includes a database with expanded instructions, a benchmark suite, and high-quality trajectories across environments. Next, we propose a novel method, \AgentEvol, to investigate the potential of agent self-evolution beyond previously seen data across tasks and environments. Experimental results show that the evolved agents can achieve results comparable to SOTA models. 
We release the \AgentGym suite, including the platform, dataset, benchmark, checkpoints, and algorithm implementations.
\\
\\
Project site: \href{https://agentgym.github.io}{https://agentgym.github.io} \\
\AgentGym suite: \href{https://github.com/WooooDyy/AgentGym}{https://github.com/WooooDyy/AgentGym}

\end{abstract}

\section{Introduction}
\begin{figure*}[t]
    \includegraphics[width=0.99\linewidth]{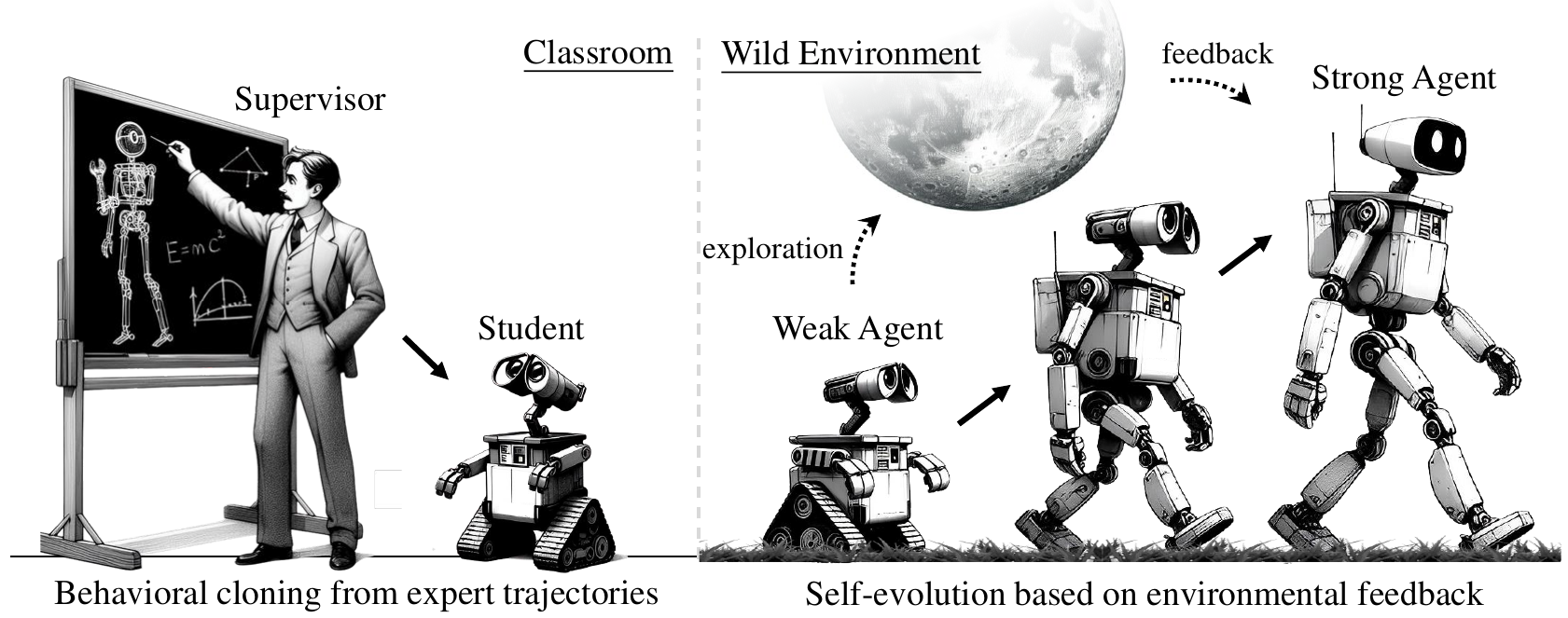}
    \centering
    \vspace{-0.3cm}
 	\caption{
   An illustration of self-evolution for generally-capable LLM-based agents in our paper. The agent first performs behavioral cloning according to human supervision, and then performs exploring and learning across environments and tasks to evolve themselves.
  }\label{fig:main}
  \vspace{-0.18cm}
\end{figure*}

Developing agents capable of performing a wide spectrum of tasks across diverse environments at human-level has been a long-standing goal for AI community, and significant endeavors have been undertaken \citep{DBLP:journals/ker/WooldridgeJ95,DBLP:journals/nature/SilverSSAHGHBLB17,silver2018general,DBLP:journals/tmlr/ReedZPCNBGSKSEBREHCHVBF22,DBLP:journals/corr/abs-2309-07864}. Similar to human learning, an agent starts by acquiring basic knowledge and skills through imitation \citep{DBLP:journals/tmlr/ReedZPCNBGSKSEBREHCHVBF22,DBLP:conf/nips/FanWJMYZTHZA22}. As it progresses, the agent is expected to continuously learn and adapt to previously unseen tasks by interacting with different environments \citep{standish2003open,DBLP:journals/alife/TaylorBCABBDFHI16,DBLP:conf/nips/FanWJMYZTHZA22,DBLP:journals/corr/abs-2404-10179}. Moreover, it may harness a rich source of insights and wisdom from its own experiences as well as those of others, developing a certain level of generalization ability \citep{tao2024survey,bran2023chemcrow}. Figure \ref{fig:main} illustrates this evolution process.

Large language models (LLMs) are considered a promising foundation for constructing such generalist agents due to their generalized abilities \citep{DBLP:journals/corr/abs-2303-08774,Claude3,DBLP:journals/corr/abs-2312-11805}, and many efforts have been made in this realm \citep{DBLP:journals/corr/abs-2309-07864,DBLP:journals/fcsc/WangMFZYZCTCLZWW24}. 
One line of work primarily relies on human supervision, where LLM-based agents mimic expert-provided trajectories from various environments step-by-step, i.e. behavioral cloning (BC) \citep{zeng2023agenttuning,DBLP:journals/corr/abs-2403-12881,zhang2024agentohana}. This method, while effective, requires skilled annotators and significant financial resources, making it hard to scale \citep{DBLP:journals/corr/abs-2403-14589}. 
Moreover, such a paradigm may encounter bottlenecks in performance, adaptability, and generalization due to insufficient exploration of the environments \citep{DBLP:journals/corr/abs-2312-10003}. 
Another line of research allows LLM-based agents to improve themselves based on environmental feedback (i.e., self-improvement), reducing reliance on human supervision while enriching exploration of the environment \citep{DBLP:journals/corr/abs-2402-19446,tao2024survey,DBLP:journals/corr/abs-2403-02502}. 
Yet, they typically train agents in isolated environments on specific tasks such as web navigation, and these specialist agents can not generalize beyond narrow tasks.

In this paper, we take the initial step to investigate the potential of self-evolution in generally capable LLM-based agents across various tasks and environments, progressing from imitation to interactive learning, akin to humans (Figure \ref{fig:main}).
% In this paper, we investigate the potential of self-evolution for generally-capable LLM-based agents across various tasks and environments (Figure \ref{fig:main}), which we believe is of significant importance to achieve human-level intelligence.
We identify three key pillars necessary for this research goal.
First, diverse environments and tasks that allow the agents to evolve dynamically and comprehensively, rather than being confined to an isolated world, which may limit generalization \citep{standish2003open,langdon2005pfeiffer,DBLP:journals/alife/TaylorBCABBDFHI16,DBLP:conf/nips/FanWJMYZTHZA22}. 
Second, a trajectory set of an appropriate size to train a base agent with preliminary instruction-following abilities and knowledge. This facilitates further exploration as in diverse, complex environments, it would be extremely inefficient for an agent to learn everything from scratch through trial and error \citep{DBLP:conf/nips/FanWJMYZTHZA22,DBLP:journals/corr/abs-2403-02502}. 
Third, an effective and flexible evolution method can adapt to environments of varying difficulty and elicit the generalizing ability of LLM-based agents. This involves how the agent interacts with the environment and how it utilizes the feedback \citep{DBLP:journals/corr/abs-2403-14589,DBLP:journals/corr/abs-2312-10003}.

\begin{figure}[t]
    \includegraphics[width=0.999\linewidth]{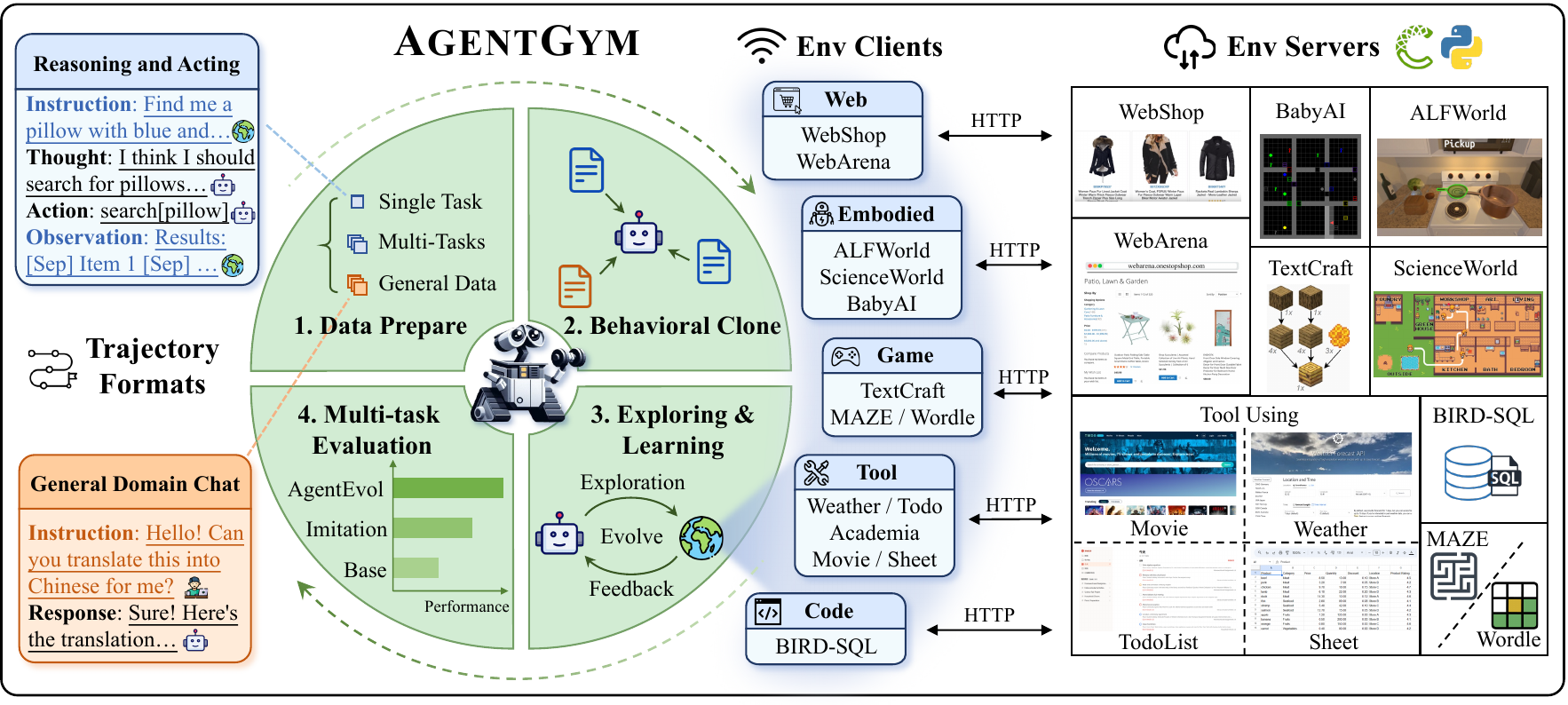}
    \centering
    \vspace{-0.3cm}
 	\caption{
  Overview of the \AgentGym framework. It covers fourteen environments spanning diverse categories.
  Each environment is deployed as an HTTP service, and clients provide encapsulated, unified interfaces for agents, facilitating interaction with environments. We gather expert-annotated trajectories from diverse environments, called \AgentTraj. We then let the agent perform behavioral cloning on this set to obtain a base generally-capable agent. With our \AgentEvol method, we explore the agent's evolution across various environments and tasks. Finally, we evaluate the agent comprehensively using the proposed benchmark suite \AgentEval.
  }\label{fig:agentgym}
  \vspace{-0.1cm}
\end{figure}

Considering the three pillars, we present \AgentGym (see Figure \ref{fig:agentgym}), a new framework designed to help the community develop generally-capable LLM-based agents and explore self-evolution. Our main contributions are:

\textbf{1. An interactive platform that includes diverse environments, tasks, and goals for LLM-based agents.} \AgentGym offers convenient APIs through HTTP services, standardizing task specifications, environment settings, and the observation/action spaces for agents. 
Within this platform, we have implemented a unified interface for multi-round interactions and real-time feedback across different environments to support online evaluation, trajectory sampling, and interactive training. 
Specifically, it includes $14$ types of agent environments, $89$ types of tasks, spanning web tasks \citep{DBLP:journals/corr/abs-2307-13854,DBLP:conf/nips/Yao0YN22}, embodied tasks \citep{DBLP:conf/emnlp/WangJCA22,DBLP:conf/iclr/Chevalier-Boisvert19}, and more \citep{abdulhai2023lmrl,DBLP:conf/iclr/ShridharYCBTH21,DBLP:journals/corr/abs-2311-05772,DBLP:journals/corr/abs-2401-13178,DBLP:conf/nips/ZhengC00WZL0LXZ23}, with high flexibility to expand to additional ones.

\textbf{2. Expanded instructions, benchmark suite, and high-quality interactive trajectories across environments. }
We collect instructions from various environments and tasks, expanding them through crowdsourcing and AI-based methods such as self-instruct \citep{DBLP:conf/acl/WangKMLSKH23} and instruction evolution \citep{DBLP:journals/corr/abs-2304-12244}. Subsequently, we select a diverse and challenging subset to form the test set to construct a benchmark suite named \AgentEval. 
Next, using crowdsourcing procedures and state-of-the-art (SOTA) models, we annotate and filter a trajectory set in a uniform format named \AgentTraj. This set is used to train a base generally-capable agent, bootstrapping further agent exploration and evolution.
For a fair comparison, we also collect a larger trajectory set \AgentTraj-L with the same pipeline to train an agent that serves as the maximum performance achievable through BC. Note that \AgentTraj-L is an extension of \AgentTraj.

\textbf{3. Initial investigation of self-evolution for generally-capable LLM-based agents based on environmental feedback.} 
Starting from the base generally-capable agent, we propose \AgentEvol, a novel method to explore agent evolution
across multiple environments.
Our focus is on investigating whether agents can evolve themselves when facing previously unseen tasks and instructions, which requires them to perform exploration and learn from new experiences. 
Experimental results show that the evolution of agents is very pronounced, even achieving comparable or better performance than SOTA models. Moreover, we perform sufficient ablation and analysis to show how our method works.

In summary, we present \AgentGym, a new framework that includes an interactive platform of multiple agent environments, a benchmark suite \AgentEval, and two trajectory sets \AgentTraj and \AgentTraj-L. We also propose a new algorithm \AgentEvol to explore self-evolution in generally-capable LLM-based agents. We will release the whole suite, algorithm implementations, and agent checkpoints. 
We hope that \AgentGym will help the community to develop new algorithms and advancements towards better generalist LLM-based agents.

\section{Preliminaries}\label{sec:Preliminaries}

We define the collection of environments as $\mathcal{E}$. For a specific ${e} \in \mathcal{E}$, we formalize the agent task in the environment as a partially observable Markov decision process (POMDP) $(\mathcal{U},\mathcal{S},\mathcal{A},\mathcal{O},\mathcal{T},r)_{e}$ with instruction space $\mathcal{U}$, state space $\mathcal{S}$, action space $\mathcal{A}$, observation space $\mathcal{O}$, deterministic state transition function $\mathcal{T}: \mathcal{S} \times \mathcal{A} \rightarrow \mathcal{S}$, and reward function ${r}: \mathcal{S} \times \mathcal{A} \rightarrow \mathbb{R}$.

Given a task instruction ${u}$ in environment $e$, the LLM-based agent parameterized by $\theta$ generates an action $a_1 \sim \pi_{\theta}(\cdot|e,u)$ based on its policy $\pi_\theta$. Then, the state space is transitioned to $s_1 \in \mathcal{S}$, and the agent receives feedback $o_1 \in \mathcal{O}$. 
Subsequently, the agent interacts with the environment until the task ends or exceeds the maximum number of steps. 
We adopt ReAct \citep{DBLP:conf/iclr/YaoZYDSN023} to modeling agent outputs, where the LLM-based agent generates a reasoning thought before outputting an action.
Thus, at time step $t$, given the history and current feedback, the agent generates the thought $h_{t+1} \sim \pi_{\theta}(\cdot|e,u,h_1,a_1,o_1,...,h_{t},a_{t},o_{t})$ first and the subsequent action $a_{t+1} \sim \pi_{\theta}(\cdot|e,u,h_1,a_1,o_1,...,h_{t},a_{t},o_{t},h_{t+1})$. The trajectory is represented as:
\begin{align}
\tau = (h_1,a_1,o_1,...,o_{T-1},h_{T},a_{T}) \sim \pi_{\theta}(\tau|e,u), 
\end{align}
\begin{equation}
\begin{aligned} 
 \pi_{\theta}(\tau|e,u) &= \prod_{t=1}^{T} \pi_{\theta}(h_t, a_t|e,u,c_{t-1}) = \prod_{t=1}^{T} \pi_{\theta}( a_t|e,u,c_{t-1},h_t) \cdot \pi_{\theta}(h_t|e,u,c_{t-1}),
\end{aligned}
\end{equation}
where $T$ is the interaction rounds, and $c_{t-1}=(h_1,a_1,o_1,...,h_{t-1},a_{t-1},o_{t-1})$ represents the interactive history up to $t-1$. The final reward $r(e,u,\tau) \in [0,1]$ is then computed.

\section{\AgentGym: Platform, Benchmark Suite and Trajectory Set}\label{sec:AgentGym_framework}

\begin{wraptable}[29]{r}{0.5\textwidth}
    \begin{minipage}{0.5\textwidth}
    \centering
    \vspace{-20pt}
        \caption{
         Comparison of \AgentGym with other agent frameworks covers several aspects: the number of environments, presence of an interactive platform and its usage, availability of trajectory sets, support for evolution, and the evolution mode.
        }
        \resizebox{\linewidth}{!}{
        \begin{tabular}{lcccc}
        \toprule
        Frameworks & Env. & Inter. Plat. & Traj. & Evol. \\
        \midrule
        AgentBench \citep{DBLP:journals/corr/abs-2308-03688}  & $8$ &  Eval   & \textcolor[HTML]{c92a2a}{No}  & \textcolor[HTML]{c92a2a}{No} \\
        AgentBoard \cite{DBLP:journals/corr/abs-2401-13178} & $12$  & Eval  & \textcolor[HTML]{c92a2a}{No}  & \textcolor[HTML]{c92a2a}{No} \\
        AgentOhana  \citep{zhang2024agentohana}  & $10$ & \textcolor[HTML]{c92a2a}{No} & Yes  & \textcolor[HTML]{c92a2a}{No}  \\
        Pangu-Agent \cite{DBLP:journals/corr/abs-2312-14878}  & $6$  & \textcolor[HTML]{c92a2a}{No} & Yes &  Single-Env  \\
        \AgentGym (Ours)& $14$ & Eval \& Train & Yes &  Multi-Env \\
        \midrule
        \end{tabular}
        }
        \label{table:different methods comparison}
    \end{minipage}
    \begin{minipage}{0.5\textwidth}
    \centering
    % \vspace{-18pt}
    \caption{Statistics of \AgentGym, including the count of task types, instruction set size, evaluation set size, size of trajectory sets (\AgentTraj and \AgentTraj-L), and average rounds of each environment.}
    \resizebox{\linewidth}{!}{
    \begin{tabular}{lcccccc}
    \toprule
    Env. &Task Num & Instr. &Eval. &Traj & Traj-L  &Rounds\\
    \midrule
    WA & $3$ & $812$ & $20$ & $0$ & {$0$}  & $-$  \\
    WS & $1$ & $6910$ & $200$ & $1000$ & {$3930$} & $5.1$  \\
    MZ & $1$ & $240$ & $25$ & $100$ & {$215$}  & $4.3$  \\
    WD & $1$ & $980$ & $25$ & $500$ & {$955$}  & $4.3$  \\
    ALF & $6$ & $3827$ & $200$ & $500$ & {$2420$} & $13.3$  \\
    Sci & $30$ & $2320$ & $200$ & $1000$ & {$2120$} & $19.9$  \\
    Baby & $40$ & $900$ & $90$ & $400$ & {$810$}  & $5.7$  \\
    TC & $1$ & $544$ & $100$ & $300$ & {$374$}  & $8.0$    \\
    WT & $1$ & $343$ & $20$ & $160$ & {$311$}  & $5.5$   \\
    MV & $1$ & $238$ & $20$ & $100$ & {$215$}  & $4.0$   \\
    AM & $1$ & $20$ & $20$ & $0$ & {$0$}  & $-$   \\
    ST & $1$ & $20$ & $20$ & $0$ & {$0$}  & $-$  \\
    TL & $1$ & $155$ & $20$ & $70$ & {$135$}  & $5.6$  \\ 
    BD & $1$ & $3200$ & $200$ & $2000$ & {$3000$}  & $1.0$   \\
   
    \midrule
    Total & $89$ & $20509$ & $1160$ & $6130$ & $14485$  & $-$  \\
    
    \bottomrule
    \end{tabular}
    }
    \label{tab:env_statistics}
    \end{minipage}
\end{wraptable}

\AgentGym is a framework designed to help the community easily evaluate and develop generally-capable LLM-based agents. It features diverse interactive environments and tasks with a unified format, i.e., ReAct format \citep{DBLP:conf/iclr/YaoZYDSN023}. It supports real-time feedback and concurrency, and is easily scalable. 
We include $14$ environments and $89$ tasks across web navigating, text games, house-holding tasks, digital games, embodied tasks, tool-using and programming. They are challenging for current LLM-based agents.
For web navigating task, we introduce WebArena (WA) \citep{DBLP:journals/corr/abs-2307-13854} and WebShop (WS) \citep{DBLP:conf/nips/Yao0YN22}. We include MAZE (MZ) and Wordle (WD) \citep{abdulhai2023lmrl} in text games. We choose ALFWorld (ALF) \citep{DBLP:conf/iclr/ShridharYCBTH21} for house-holding tasks. We include SciWorld (Sci) \citep{DBLP:conf/emnlp/WangJCA22} and BabyAI (Baby) \citep{DBLP:conf/iclr/Chevalier-Boisvert19} in embodied tasks. We choose TextCraft (TC) \citep{DBLP:journals/corr/abs-2311-05772} for digital games. We get Tool-Weather (WT), Tool-Movie (MV), Tool-Academia (AM), Tool-Sheet (ST) and Tool-TODOList (TL) \citep{DBLP:journals/corr/abs-2401-13178}  for tool using tasks. We establish BIRD (BD) \citep{DBLP:conf/nips/ZhengC00WZL0LXZ23} for programming tasks. See Appendix \ref{sec: env details} for environment details.
% $\bigcup\limits_{}^{}$
The comparison between \AgentGym and other frameworks is demonstrated in Table \ref{table:different methods comparison}.
\paragraph{Platform architecture and components.}
Recognizing the diverse dependencies inherent to different agent environments, \AgentGym deploys separate services for each environment in a user-friendly manner to prevent conflicts. The clients can communicate with environments using HTTP protocol.  At the core of this architecture is the controller, which acts as a conduit for interactions between agents and environmental services, providing an encapsulated, unified interface of environmental functions or operations for agents to invoke. Additionally, we implement user-friendly components such as the evaluator, trainer, and data collection pipeline to support community development. Figure \ref{fig:platform_arch} in Appendix \ref{appendix:platform_arch} illustrate the architecture design of the platform.

\paragraph{Instruction collecting and benchmark construction.}
We have collected $20509$ instructions and queries across environments and tasks. 
For tasks that already have an abundance of instructions, such as WebShop and ALFWorld, we primarily rely on their original source. Meanwhile, for tasks with fewer instructions, like tool-using tasks, we expand upon them using self-instruct and instruction evolution methods, specifically by prompting GPT-4 to generate new instructions \citep{DBLP:conf/acl/WangKMLSKH23,DBLP:journals/corr/abs-2304-12244}.
The details are in Appendix \ref{sec: env details}.
We then extract a diverse and challenging subset $\mathcal{Q}_{eval}$ of $1160$ instructions from each environment to construct a benchmark suite \AgentEval, which can comprehensively evaluate LLM-based agents. The remained instruction set is denoted as $\mathcal{Q}=\bigcup_{e \in \mathcal{E}}\mathcal{Q}_e$, where $\mathcal{Q}_e$ means the remained instructions of environment $e$.
\paragraph{Trajectory collecting and filtering.}
In \AgentGym, the server provides instructions including task description, environment setup, and problem to the agent.
Next, as described in Section \ref{sec:Preliminaries}, the agent interacts with the environment in ReAct-Style until the task is completed. 
We collect trajectories with SOTA models (e.g., GPT-4-Turbo) and crowdsourced annotations. Details are in Appendix \ref{sec: env details}. We rigorously filter the trajectories to ensure data quality based on the reward or correctness, and get a set of $6130$ trajectories. This set, named \AgentTraj, is used to train a base generally-capable agent in Section \ref{sec:Behavioral Cloning with Collected Trajectories}. 
For a fair comparison, we perform annotation and filtering on all instructions using the same pipeline and get \AgentTraj-L to show the performance upper bound of BC.

Detailed statistics of \AgentGym framework are shown in Table \ref{tab:env_statistics}. 

\section{\AgentEvol for Evolution of Generally-capable LLM-based Agents} \label{sec:AgentEvol for Evolution of Generally-capable LLM-based Agents}
In this section, we first train a base generally-capable agent through behavioral cloning to equip it with basic interactive ability in agent tasks. Building on this agent, we take the initial steps to explore the comprehensive evolution of LLM-based agents across multiple environments and tasks. We summarize the algorithm in Algorithm \ref{Algorithm: AgentEvol}.
\subsection{Behavioral Cloning with Collected Trajectories}\label{sec:Behavioral Cloning with Collected Trajectories}

Behavioral cloning fine-tunes LLM-based agents by having them mimic the collected expert trajectories step-by-step. In practice, we expect the agent to accomplish the appropriate inner thought $h$ and action $a$. 
We use \AgentTraj (denoted as $\mathcal{D}_s$) to train a base generally-capable agent with basic instruction-following ability and prior knowledge. 
% The statistics of $\mathcal{D}_{sub}$ are shown in the Traj-S column of Table \ref{tab:env_statistics}. 
We maximize the following objective:
\begin{equation}
\begin{aligned} 
    \mathcal{J}_{BC}(\theta) &=  \mathbb{E}_{(e,u,\tau) \sim \mathcal{D}_{s}}\Big [\log{\pi_{\theta}}(\tau|e,u)\Big ] \\
     &= \mathbb{E}_{(e,u,\tau) \sim \mathcal{D}_{s}}{\sum_{t=1}^T \Big [\log \pi_{\theta}(a_t|e,u,c_{t-1},h_t) + \log \pi_{\theta}(h_t|e,u,c_{t-1}) } \Big ].
    \label{eqn:J_BC}
\end{aligned}
\end{equation}
Note that we include a general domain dataset $\mathcal{D}_{general}$ as in Zeng et al. \citep{DBLP:journals/corr/abs-2310-12823} to maintain the agent's ability in language understanding and generation. And the resulting agent $\pi_{\theta_{base}}$ serves as a starting point for later evolution across diverse environments and tasks.

\subsection{Evolution through Exploration and Learning}

This work tries to explore the potential of self-evolution in generally-capable LLM-based agents across multiple environments and tasks. 
\textbf{More importantly, the agents will face previously unseen tasks and instructions during evolution.}
Hence, agents are required to explore environments, receive feedback, and optimize themselves based on the feedback. 
To achieve our goal, reinforcement learning (RL) \citep{sutton2018reinforcement} is worth considering, and the corresponding objective is:
\begin{equation}
\begin{aligned} 
   \mathcal{J}_{RL}(\theta) = \mathbb{E}_{e \in \mathcal{E},u \sim \mathcal{Q}_e,\tau \sim \pi_{\theta}(\tau | e,u)}[r(e,u,\tau)]  .
   \label{eqn:J_RL}
\end{aligned}
\end{equation}
However, in our setting, standard RL faces significant challenges due to large sampling space and long-term nature of agent tasks, leading to high computational complexity and training instability, which hampers scalability \citep{huang202237,DBLP:journals/corr/abs-2307-04964,DBLP:journals/corr/abs-2402-05808}. 
Hence, we draw inspiration from the well-established connection between RL and probabilistic inference \citep{dayan1997using,DBLP:conf/icml/Neumann11,DBLP:conf/ijcai/RawlikTV13a,DBLP:conf/iclr/AbdolmalekiSTMH18}, and propose a method called \AgentEvol for agent evolution, which involves agents alternating between exploration and learning.
\paragraph{Learning from estimated optimal policy.}
 
In this work, we view RL as an inference problem within a specific probabilistic model \citep{dayan1997using,DBLP:conf/iclr/AbdolmalekiSTMH18,DBLP:journals/corr/abs-2312-06585}. Differing from traditional RL formulations that focus on identifying a trajectory that maximizes the expected reward, inference-based approaches start with an optimal distribution over trajectories. 
We initially define $P( O=1 )$ to represent the event of ``obtained optimal policy by maximum expected rewards'' or ``achieving success in the RL task'', which can be calculated by integrating the optimal policy probability at each sampling point. 
Given the policy agent $\pi_\theta$, the optimal policy can be obtained by maximizing:
\begin{equation}
\label{eq:obj}
\begin{aligned}
    \log P_{\pi_\theta}(O = 1) &= \log \int \pi_\theta(\tau)p(O = 1|\tau)d\tau  .
\end{aligned}
\end{equation}

However, the above optimization process is difficult to proceed directly due to the fact that LLM-based agents require token-wise feedback to perform gradient updates. In this paper, we alternatively construct the variational lower bound of Eq.\ref{eq:obj} by introducing an estimation function $q$ on the optimal policy. With Jensen's inequality, we soon have:
\begin{equation}
\begin{aligned}
\log \int \pi_\theta(\tau)p(O = 1|\tau)d\tau  =& \log  \mathbb{E}_{q(\tau)}[\frac{\pi_\theta(\tau)}{q(\tau)} p(O = 1|\tau)]
    \geq \mathbb{E}_q [\log \frac{\pi_\theta(\tau)}{q(\tau)} p(O = 1|\tau)] \\
    =& \ \mathbb{E}_q [\log p(O=1|\tau)] - \text{KL}[q(\tau)||\pi_\theta(\tau)] = \mathcal{J}(q,\pi_\theta),
    \label{eqn:J_q_pi}
\end{aligned}
\end{equation}
where $\pi_\theta$ is the trajectory distribution induced by the agent, and $q(\tau)$ is a variational distribution.

Due to the monotonicity of the logarithmic function, by maximizing the lower bound $\mathcal{J}(q, \pi_{\theta})$, we can obtain a policy with an expected return higher than before. Generally, our framework can be divided into two steps of loop iteration. 
The former step of $\mathcal{J}(q, \pi_{\theta})$ can be explained as estimating the optimal policy distribution on the sampled trajectories by maximizing the expected reward over the state space. The later step relates to updating the current agent's parameters $\theta$ towards the optimal policy $q$, thus completing the optimization of one single iteration. In analogy to SGD \citep{robbins1951stochastic}, the estimation process introduces noise to the policy optimization due to the presence of unseen decision trajectories. This error gradually decreases as the algorithm proceeds and converges to zero when the current agent becomes optimal \citep{dayan1997using}.

In the \AgentEvol algorithm, we refer to the two steps as \textbf{Exploration step} and \textbf{Learning step}. Specifically, with current agent parameters \(\theta^m\) and the variational distribution \(q^m\) \citep{DBLP:journals/corr/abs-2312-06585}, at \textbf{exploration step}, the estimation of optimal policy $q$ is updated by maximizing the expected reward: $q^{m+1}=\arg \max_{q} \mathcal{J}(q, \pi_{\theta^m})$. As $\max_q \mathcal{J}(q,\pi_{\theta^{m}}) = \min_q[\text{KL}(q(\tau)\|p(O=1|\tau)\pi_{\theta^m}(\tau))]$, we have $q^{m+1} \propto p(O=1|\tau)\pi_{\theta^m}(\tau)$.
This step is equivalent to evaluating the likelihood that the samples generated from the current agent's policy achieve best rewards, and observe the returns of $q$ by empirically estimating on a pre-constructed training set.
And at \textbf{learning step}, we optimizes \(\mathcal{J}(q^{m+1}, \pi_{\theta})\) by updating \(\theta\). This process is similar to learning a new distribution sampled from the optimal policy on the original training data. Since the first term of \(\mathcal{J}(q^{m+1}, \pi_{\theta})\) relates only to $q$ as well as $\tau$, the training objective is equivalent to measuring the KL divergence between the estimated policy $q^{m+1}(\cdot)$ and the current policy $\pi_\theta (\cdot)$ over all training samples. We finally derive:
\begin{equation}
    \begin{aligned}
        \theta^{m+1} \coloneqq& \arg \min_\theta \text{KL}[q^{m+1}(\tau) \| \pi_\theta (\tau)] \\
        =& \arg \min_\theta \sum_\tau -q^{m+1}(\tau) \log \pi_\theta (\tau).
    \end{aligned}
\end{equation}
This involves optimizing a weighted negative log-likelihood function based on \( q^{m+1} \), which adjusts the agent policy to increase the likelihood of generating higher-reward trajectories, thereby improving the agent's performance.

\paragraph{Practical evolution for LLM-based agent.} 
\begin{algorithm}[t]
\caption{\AgentEvol}
\label{Algorithm: AgentEvol}
  \SetKwData{Left}{left}\SetKwData{This}{this}\SetKwData{Up}{up}
  \SetKwFunction{Union}{Union}\SetKwFunction{FindCompress}{FindCompress}
  \SetKwInOut{Input}{Input}\SetKwInOut{Output}{Output}
   \SetKwProg{myproc}{Procedure}{}{}
   \KwIn{Initialized policy LLM-based agent $\pi_\theta$, environment set $\mathcal{E}$, trajectory subset $\mathcal{D}_{s}$, full instruction set $\mathcal{Q}$, reward function $r$.}
   \myproc{{{\textnormal{Behavioral cloning:}}}}{  
        Maximize objective $\mathcal{J}_{BC}(\theta) =  \mathbb{E}_{(e,u,\tau) \sim \mathcal{D}_{s}}\Big [\log{\pi_{\theta}}(\tau|e,u)\Big ]$ to get $\pi_{\theta_{base}}$;  \\
    }
  \myproc{{{\textnormal{Evolution :}}}}{  

   $\pi_{\theta^1} \gets \pi_{\theta_{base}}$; \\
        \For{\ \textnormal{iteration} \  $m=1$ \textnormal{to} $M$ \ \ }{
            \textcolor{violet!70!red}{// Perform \textbf{Exploration Step}}\\
            $\mathcal{D}_m = \bigcup_{e \in \mathcal{E}} \mathcal{D}_m^e$, where $\mathcal{D}_m^e = \{(e,u^j,\tau^j)\ | u^j \sim \mathcal{Q}_e, \tau^j \sim \pi_{\theta^m}(\tau|e,u^j)\}_{j=1}^{|\mathcal{D}_m^e|}$; \\
            Compute reward for  $\mathcal{D}_m$ with $r$; \\
            $\mathcal{D}_m  \gets \mathcal{D}_m \cup \mathcal{D}_{s}$;\\
            \textcolor{red!30!cyan}{// Perform \textbf{Learning Step}} \\ 
            Maximize objective $\mathcal{J}_{Evol}(\theta) = \mathbb{E}_{(e,u,\tau) \sim \mathcal{D}_m}[r(e,u,\tau)\log \pi_\theta (\tau|e,u)]$ to get $\pi_{\theta^{m+1}}$;
        }
    }
\end{algorithm}

In our LLM-based agent scenario, the trajectory is conditioned on the environment \(e\) and instruction \(u\). Considering our non-negative reward function \(r(e,u,\tau)\), we can get \(P(O|e,u,\tau) \propto r(e,u,\tau)\) \citep{DBLP:journals/corr/abs-2312-06585}. Consequently, \(q^{m+1} (\tau|e,u) \propto r(e,u,\tau) \cdot \pi_{\theta^{m}} (\tau|e,u)\). Thus, the policy update in the learning is:
\begin{equation}
    \begin{aligned}
        \theta^{m+1} \coloneqq& \arg \min_\theta \sum_\tau - (r(e,u,\tau) \cdot \pi_\theta (\tau|e,u)) \log \pi_\theta (\tau|e,u) \\
        =& \arg \max_\theta  \mathbb{E}_{e \in \mathcal{E} ,u \sim \mathcal{Q}_e,\tau \sim \pi_{\theta^m}(\tau | e,u)}[r(e,u,\tau) \log \pi_\theta (\tau|e,u) ] .
    \end{aligned}
\end{equation}
This can be viewed as a supervised learning objective weighted by the reward. This approach uses the fixed policy agent from the previous iteration to sample data, thereby separating data collection and policy optimization. In contrast, standard RL performs on-policy data sampling and optimization.

Now we describe the two steps of evolution part in \AgentEvol in practice:
\begin{description}
    \item \textbf{Exploration Step.} In the $m$-th exploring iteration, for each environment $e$, we have an instruction set $\mathcal{Q}_e$ which is larger than that used in the BC phase, allowing us to investigate agents evolving to unseen tasks and instructions. 
    The current policy agent interacts with this environment, generating a collection of interaction trajectories $\mathcal{D}_m^e = \{(e,u^j,\tau^j)\ | u^j \sim \mathcal{Q}_e, \tau^j \sim \pi_{\theta^m}(\tau|e,u^j)\}_{j=1}^{|\mathcal{D}_m^e|}$. 
    Subsequently, based on the reward function of the environment, we calculate the reward $r(e,u,\tau)$ for each trajectory. The generated dataset from each environment is then merged, resulting in $\mathcal{D}_m = \bigcup_{e \in \mathcal{E}} \mathcal{D}_m^e$. Note that we also include the original trajectory set in Section \ref{sec:Behavioral Cloning with Collected Trajectories} for the learning step, i.e., $\mathcal{D}_m = \mathcal{D}_m\bigcup\mathcal{D}_{s}$.
    \item \textbf{Learning Step.} In the $m$-th learning iteration, we utilize the dataset $\mathcal{D}_m$ obtained from the exploration step to fine-tune the agent with the objective $\mathcal{J}_{Evol}(\theta) = \mathbb{E}_{(e,u,\tau) \sim \mathcal{D}_m}[r(e,u,\tau)\log \pi_\theta (\tau|e,u)]$ to get $\pi_{\theta^{m+1}}$. We also include the general domain dataset as in the BC phase.
    We optimize the initial agent $\pi_{\theta}$ at each iteration, aiming to minimize overfitting and prevent drift from the base agent. 
    In this learning step, the agent is improved, similar to previous work done on LLM reasoning \cite{DBLP:conf/nips/ZelikmanWMG22,DBLP:journals/corr/abs-2312-06585,DBLP:journals/corr/abs-2312-10003}. 
\end{description}
By alternating between the two steps, empirical results show that our method facilitates the evolution of an LLM-based agent across both seen and unseen tasks and instructions.

\section{Experiments and Discussion}

\vspace{-1mm}
\subsection{Experimental Setup}\label{sec:Experimental Setup}
\paragraph{Environments and Tasks.}
We explore the self-evolution of generally-capable LLM-based agents with the \AgentGym framework. Main experiments cover 11 environments: WebShop \citep{DBLP:conf/nips/Yao0YN22}, ALFWorld \citep{DBLP:conf/iclr/ShridharYCBTH21}, SciWorld \citep{DBLP:conf/emnlp/WangJCA22}, BabyAI \citep{DBLP:conf/iclr/Chevalier-Boisvert19}, TextCraft \citep{DBLP:journals/corr/abs-2311-05772}, BIRD \citep{DBLP:conf/nips/ZhengC00WZL0LXZ23}, MAZE, Wordle \citep{abdulhai2023lmrl}, Tool-TODOList, Tool-Weather, and Tool-Movie \citep{DBLP:journals/corr/abs-2401-13178}. Note that instructions used in BC are fewer than those in evolution, to study the agent's ability to generalize when performing exploration.
\vspace{-1mm}
\paragraph{Baselines.}
We include closed-source models like GPT-3.5-Turbo \citep{DBLP:conf/nips/Ouyang0JAWMZASR22}, GPT-4-Turbo \citep{DBLP:journals/corr/abs-2303-08774}, Claude 3 \citep{Claude3}, and DeepSeek-Chat \citep{deepseekai2024deepseekv2}. We also include open-source models like Llama-2-Chat \citep{touvron2023llama}, and agents trained on expert trajectories, i.e., AgentLM \citep{DBLP:journals/corr/abs-2310-12823}. For a fair comparison, we include a baseline that performs BC on \AgentTraj-L, serving as the maximum performance achievable through BC. 

\begin{table*}[t]
\centering
\caption{Evaluating results on diverse tasks. BC$_{base}$ means the agent trained with \AgentTraj, providing a base agent with basic ability and prior knowledge. BC$_{large}$ means the agent that performs BC on \AgentTraj-L, representing the performance upper limit of BC in this paper. It rivals, or even surpasses SOTA models and agents. 
Our evolution method, \AgentEvol, 
outperforms BC$_{large}$ on most tasks and environments through exploration and learning. The best performance of each part is highlighted in \textbf{bold}.}
\resizebox{0.9999\textwidth}{!}{ 
\begin{tabular}{lccccccccccc}
\toprule
Method &WS &ALF & TC &Sci &Baby &MZ &WD &WT &MV &TL &BD \\

\cmidrule(l){1-1} 
\cmidrule(l){2-12}
\multicolumn{12}{c}{\textbf{Close-sourced Models \& Agents}} \\ 

DeepSeek-Chat  &$11.00$ & $51.00$& $23.00$& $\mathbf{16.80}$& $45.67$& $4.00$& $24.00$& $70.00$& $70.00$& $75.00$& $13.50$ \\
Claude-3-Haiku  & $5.50$ & $0.00$& $0.00$& $0.83$& $1.93$& $4.00$& $16.00$& $55.00$& $50.00$& $65.00$& $13.50$ \\
Claude-3-Sonnet  & $1.50$ & $13.00$& $38.00$& $2.78$& $\mathbf{79.25}$& $0.00$& $36.00$& $65.00$& $80.00$& $80.00$& $\mathbf{17.00}$ \\
GPT-3.5-Turbo  & $12.50$ & $26.00$ & $47.00$ & $7.64$ & $71.36$ & $4.00$ & $20.00$ & $25.00$ & $70.00$ & $40.00$ & $12.50$   \\
 % & $20.00$ & $26.00$ & $48.00$ & $TODO$ & $63.38$ & $4.00$ & $20.00$ & $25.00$ & $70.00$ & $40.00$ & $12.50$   \\
GPT-4-Turbo  & $\mathbf{15.50}$ & $\mathbf{67.50}$ & $\mathbf{77.00}$ & $14.38$ & $72.83$ & $\mathbf{68.00}$ & $\mathbf{88.00}$ & $\mathbf{80.00}$ & $\mathbf{95.00}$ & $\mathbf{95.00}$ & $16.00$\\
% \midrule
\midrule
% \hdashline
\multicolumn{12}{c}{\textbf{Open-sourced Models \& Agents}} \\ 
Llama2-Chat-7B  &$0.50$ & $2.00$& $0.00$& $0.83$& $0.23$& $0.00$& $0.00$& $0.00$& $0.00$& $0.00$& $1.50$\\
Llama2-Chat-13B  &$1.00$ & $3.50$& $0.00$& $0.83$& $0.10$& $0.00$& $0.00$& $0.00$& $0.00$& $0.00$& $1.50$ \\
% Llama2-Chat-70B  &$TODO$ & $TODO$& $TODO$& $TODO$& $TODO$& $TODO$& $TODO$& $TODO$& $TODO$& $TODO$& $TODO$ \\
% InternLM2 &$TODO$ & $TODO$& $TODO$& $TODO$& $TODO$& $TODO$& $TODO$& $TODO$& $TODO$& $TODO$& $TODO$ \\

% \midrule
% \midrule
% \hdashline
% \multicolumn{12}{c}{\textbf{Models Fine-tuned on Agent Trajectories}} \\ 
AgentLM-7B  & $36.50$ & $71.00$ & $\mathbf{4.00}$ & $1.63$ & $0.49$ & $\mathbf{12.00}$ & $4.00$ & $0.00$ & $\mathbf{5.00}$ & $15.00$ & $5.00$  \\
AgentLM-13B  & $39.50$ & $\mathbf{73.00}$ & $0.00$ & $2.75$ & $0.45$ & $8.00$ & $0.00$ & $\mathbf{10.00}$ & $\mathbf{5.00}$ & $5.00$ & $3.00$  \\
AgentLM-70B  & $\mathbf{49.50}$ & $67.00$ & $\mathbf{4.00}$ & $\mathbf{10.68}$ & $\mathbf{0.66}$ & $8.00$ & $4.00$ & $0.00$ & $0.00$ & $\mathbf{40.00}$ & $\mathbf{7.50}$  \\
% Agent-FLAN  & $TODO$ & $TODO$& $TODO$& $TODO$& $TODO$& $TODO$& $TODO$& $TODO$& $TODO$& $TODO$& $TODO$ \\
% \midrule
\midrule
% \hdashline
\multicolumn{12}{c}{\textbf{Ours}} \\ 
BC$_{base}$ &$66.50$ & $77.50$& $44.00$& $26.42$& $69.31$& $\mathbf{12.00}$& $12.00$& $25.00$& $5.00$& $45.00$& $8.00$ \\
BC$_{large}$ & $73.50$ & $83.00$ & $60.00$ & $\mathbf{74.47}$ & $74.19$ & $\mathbf{12.00}$ & $\mathbf{36.00}$ & $\mathbf{45.00}$ & $5.00$ & $65.00$ & $8.50$  \\
% BC$_\text{\ full}$  
\textbf{\AgentEvol}  &$\mathbf{76.50}$ & $\mathbf{88.00}$& $\mathbf{64.00}$& $38.00$& $\mathbf{82.70}$& $\mathbf{12.00}$ & $12.00$& $25.00$& $\mathbf{60.00}$& $\mathbf{70.00}$& $\mathbf{9.00}$ \\
% Few-shot  & $12.3$ & $12.3$ & $12.3$ & $12.3$ & $12.3$ & $12.3$ & $12.3$ & $12.3$ & $12.3$ & $12.3$ & $12.3$ & $12.3$ \\

\bottomrule
\end{tabular}
}
\label{table:main results}
% \vspace{-5mm}
\end{table*}

\vspace{-1mm}
\paragraph{Implementation Details.}
All experiments are conducted with eight A100-80GB GPUs. Our main backbone model is Llama-2-Chat-7B. Different environments services are deployed on different ports of the same server. 
We set the iteration number $M$ to $4$.
To conserve computational resources, each instruction is sampled once during the evolution process. 
Note that some environments provide dense rewards $r \in [0,1]$, while others give only binary feedback $r \in \{0,1\}$. For simplicity and consistency, we follow previous work \citep{DBLP:journals/corr/abs-2312-06585} and use binary rewards. We set $r=0$ for trajectories where $r < 1$, while for those with $r=1$, we keep it unchanged.
See Appendix \ref{appendix:more_impl_details} for more details.

\subsection{Main Results}
Experiment results in Table \ref{table:main results} demonstrate that: (1) While closed-source models perform well, even SOTA closed-source models like GPT-4-Turbo fail to achieve satisfactory performance on all tasks, highlighting the need for developing more capable agents. 
(2) Open-source models, represented by Llama2-Chat, perform poorly on all tasks, highlighting the importance of the initialization step of BC. 
(3) Models trained on agent trajectories, like AgentLM \citep{DBLP:journals/corr/abs-2310-12823}, can perform on par with GPT-4-Turbo on many tasks, particularly the 70B version. However, they do not match performance on tasks like TextCraft \citep{DBLP:journals/corr/abs-2311-05772} or SciWorld \citep{DBLP:conf/emnlp/WangJCA22}, which can be attributed to the lack of training data. 
(4) The agent trained on \AgentTraj-L ,i.e., BC$_{large}$, achieves excellent performance, matching or even surpassing SOTA models, showing that it is a strong baseline. 
(5) \AgentEvol, despite having limited trajectories for imitation, surpasses BC$_{large}$ and SOTA models on many tasks like WebShop \citep{DBLP:conf/nips/Yao0YN22}, ALFWorld \citep{DBLP:conf/iclr/ShridharYCBTH21} and BabyAI \citep{DBLP:conf/iclr/Chevalier-Boisvert19}, validating the superiority and promise of agent evolution.

Moreover, we report the number of interactive rounds required by different models to solve the task, in order to demonstrate the efficiency of our method (Appendix \ref{appendix: Interactive Rounds}). 
% We also conduct experiments to explore evolution in isolated environments (Appendix \ref{appendix: Evolution in An Isolated Environment}) and perform case study (Appendix \ref{Appendix:case study}).

\subsection{Discussion \& Analysis} \label{sec:Discussion_and_Analysis}

\paragraph{Ablation on data merging strategies and iteration number $M$.}
\begin{wrapfigure}[20]{b}{0.38\textwidth}
  \centering
  \vspace{-10pt}
  \includegraphics[width=0.38\textwidth]{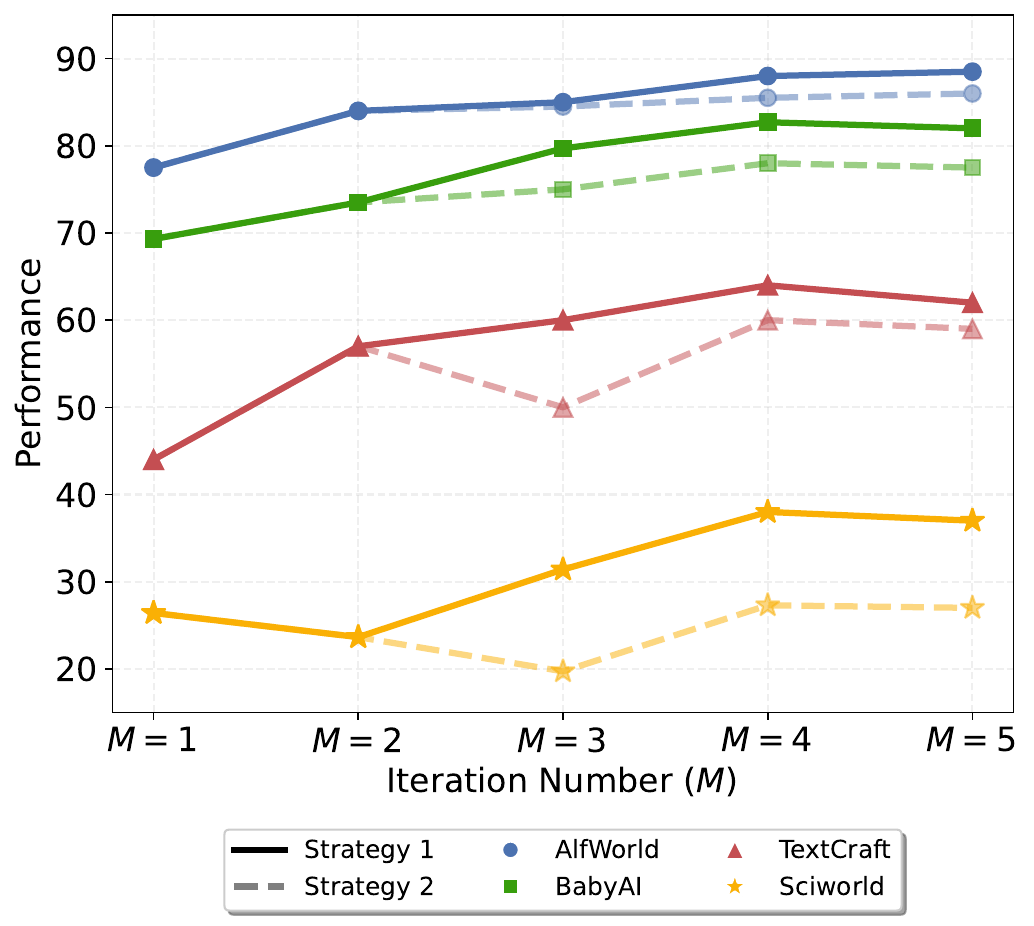}
  \vspace{-10pt}
  \caption{Ablation on data merging strategies and iteration number $M$. Strategy 1 means merging trajectories generated by the current agent with the initial trajectory set; Strategy 2 means merging current trajectories with the trajectories generated in the previous iteration.}
  \label{fig:ablation_on_data_merge_and_iter_num}
\end{wrapfigure}
In our experiments, we merge the trajectories sampled during each iteration with the initial trajectories to train the agent, rather than merging it with the trajectories generated in the previous iteration. Here, we conduct an ablation study to show the impact of this merging strategy and the iteration number $M$. 
Experimental results in Figure \ref{fig:ablation_on_data_merge_and_iter_num} show that merging with the initial data provides more stable improvements, while merging with the trajectories from the previous iteration leads to performance fluctuations, possibly due to overfitting \citep{yuan2023scaling,DBLP:journals/corr/abs-2312-06585}. 
Additionally, as $M$ increases, performance tends to improve but gradually converges in later iterations. Therefore, we choose $M=4$ to balance performance and efficiency.

\paragraph{Ablation on sample number $K$.}
In the exploration step, we perform sampling on each instruction once per iteration. Here, we conduct ablation on sample number $K$ with four tasks. The results in Table \ref{tab:ablation_on_K_and_scope} show that performance increases with higher $K$, but the improvement is not significant. So we select $K=1$ for computational efficiency.

\paragraph{Ablation on exploration scope.}
In our experiment, we first train a base agent using \( \mathcal{D}_{s} \) and then let it explore a wider range of instructions and tasks. We conduct an ablation study on four tasks to see how well the agent evolves with limited instructions as in the BC phase. Table \ref{tab:ablation_on_K_and_scope} shows that even in a limited scope, the base agent's performance improves, which may be attributed to more diverse trajectories sampled from the agent. However, the improvement is not significant, indicating that effective evolution needs a more extensive environment.

\paragraph{Effectivenes on different models.}
To demonstrate the generalizability of our method across different backbone models, we conduct experiments on Llama-2-13B \citep{touvron2023llama} and DeepSeek-Coder-1.3B \citep{guo2024deepseek}. The entire evolution process is still based on \AgentGym. The experimental results in Table \ref{tab:different_model} show that our \AgentEvol maintains its evolutionary capabilities across different backbone models, achieving performance that is comparable to or surpasses BC${_{large}}$.

\begin{table}[t]
    \centering
    \begin{minipage}[b]{0.44\textwidth}
      \centering
          % \vspace{-10pt}
          \caption{Ablation study on sample number $K$ and the exploration scope with four tasks.}
        \resizebox{\linewidth}{!}{
        \begin{tabular}{lccccc}
        \toprule
        Method &WS & ALF &Baby &TC  \\
        \midrule
        BC$_{base}$ & $66.5$ & $77.5$ & $69.3$ & $44.0$  \\
        \AgentEvol \\
        \ \ \  -w $K=1$ \  & $77.0$ & $88.0$ & $82.9$ &$65.0$ \\
        \ \ \  -w $K=2$ \  & $76.0$ & $88.0$ & $83.1$ & $67.0$  \\
        \ \ \  -w $K=3$ \  & $\mathbf{78.5}$ & $\mathbf{89.0}$ & $\mathbf{83.6}$ & $\mathbf{68.0}$ \\
        % \hdashline
        \ \ \ -w Limited Scope for Exploration & $70.0$ & $80.5$ & $70.7$ & $49.0$  \\
        
        \bottomrule
        \end{tabular}
        }
        \label{tab:ablation_on_K_and_scope}
        % \end{center}
        
    \end{minipage}
    \hfill
    \begin{minipage}[b]{0.47\textwidth}
        \centering
        % \vspace{-15pt}
        \caption{Effectiveness of our method on different models. }
        \vspace{1pt}
        \resizebox{\linewidth}{!}{
        \begin{tabular}{llccccc}
        \toprule
        Model &Method &WS & ALF &Baby &TC  \\
        \midrule
        % DeepSeek-Coder-1.3B  \\
        DeepSeek-Coder-1.3B &BC$_{base}$ & $54.0$ & $33.0$ & $68.9$ & $31.0$  \\
        & BC$_{large}$ & $65.0$ & $\mathbf{62.5}$ & $73.8$ & $37.0$  \\
        &\AgentEvol & $\mathbf{67.5}$ & $54.5$ & $\mathbf{77.3}$ & $\mathbf{38.0}$  \\
        \hdashline
        % Llama2-Chat-13B  \\
        Llama2-Chat-13B &BC$_{base}$ & $65.5$ & $81.5$ & $76.6$ & $59.0$  \\
        & BC$_{large}$ & $74.0$ & $85.0$ & $81.1$ & $61.0$  \\
        & \AgentEvol & $\mathbf{78.5}$ & $\mathbf{89.5}$ & $\mathbf{86.8}$ & $\mathbf{71.0}$  \\
        \bottomrule
        \end{tabular}
        }
        \label{tab:different_model}
        % \end{center}
        
    \end{minipage}
\end{table}

\paragraph{Evolution with both successful and failed trajectories. }
\begin{wraptable}[8]{r}{0.46\textwidth}
\centering
  \vspace{-20pt}
  \caption{Experiments on evolution with both successful and failed trajectories. }
\resizebox{\linewidth}{!}{
\begin{tabular}{lccccc}
\toprule
Method &WS & ALF &Baby &TC  \\
\midrule
BC$_{base}$ & $66.5$ & $77.5$ & $69.3$ & $44.0$  \\
\AgentEvol & $\textbf{77.0}$ & $\textbf{88.0}$ & $\textbf{82.9}$ & $\textbf{65.0}$  \\ 
DPO with failed traj & $75.0$ & $86.5$ & $78.3$ & $58.0$  \\

\bottomrule
\end{tabular}
}
\label{tab:with_failed_traj}
% \end{center}
\end{wraptable}
In learning step, we only utilize the sampled trajectories with high rewards (success) and do not use failed trajectories. Inspired by previous work \citep{DBLP:journals/corr/abs-2402-14830,DBLP:journals/corr/abs-2402-06457,DBLP:journals/corr/abs-2403-02502,DBLP:journals/corr/abs-2403-14589,wang2023making}, we explore whether failed trajectories can be included for better evolution. Specifically, we construct pairs of successful and failed trajectories and optimize the agent using the DPO method \citep{DBLP:conf/nips/RafailovSMMEF23}, which fits models to the pair-wise dataset \citep{wang2023making,DBLP:journals/corr/abs-2402-14830,DBLP:journals/corr/abs-2404-03648}. 
Results in Table \ref{tab:with_failed_traj} show that using both types of trajectories can still bring about evolutionary effects, but the performance is not as good as our method, indicating that preference optimization in multi-task setting is more challenging compared to single-task \citep{DBLP:journals/corr/abs-2402-14830,DBLP:journals/corr/abs-2403-02502}. In the future, we hope to explore more algorithms to make full use of all trajectories for comprehensive evolution.

\section{Related Work}

With the development of LLMs \citep{DBLP:journals/corr/abs-2303-08774,Claude3,DBLP:journals/corr/abs-2312-11805}, developing agents based on them has become an important research direction \citep{DBLP:journals/corr/abs-2309-07864,DBLP:journals/fcsc/WangMFZYZCTCLZWW24}. These agents are typically endowed with reasoning and acting capabilities and can perform many types of tasks \citep{DBLP:conf/iclr/YaoZYDSN023,DBLP:journals/corr/abs-2312-10003,DBLP:journals/corr/abs-2310-05915}. 
To evaluate these agents, researchers have proposed benchmarks that include various tasks \citep{DBLP:conf/nips/Yao0YN22,DBLP:journals/corr/abs-2308-03688,DBLP:journals/corr/abs-2401-13178,DBLP:journals/corr/abs-2310-11667}. 
Our benchmark suite \AgentEval offers a more diverse set of environments and tasks, providing a more comprehensive evaluation.

Closed-source LLMs, equipped with prompting methods like ReAct \citep{DBLP:conf/iclr/YaoZYDSN023} and PlanAct \citep{DBLP:journals/corr/abs-2308-05960}, can achieve great performance in agent tasks, while agents based on open-source methods perform poor on these tasks \citep{DBLP:journals/corr/abs-2308-03688,DBLP:journals/corr/abs-2312-14878}.
To address this challenge, a series of work collects expert trajectories from diverse environments and tasks and trains LLM-based agents through behavioral cloning \citep{DBLP:journals/corr/abs-2310-12823,DBLP:journals/corr/abs-2310-05915,DBLP:journals/corr/abs-2403-12881,zhang2024agentohana}. However, obtaining these expert trajectories is often costly and they lack sufficient exploration of the environment \citep{DBLP:journals/corr/abs-2403-14589,DBLP:journals/corr/abs-2312-10003}. 

Another line of work trains LLM-based agents based on environmental feedback, referred to as interactive learning methods \citep{DBLP:journals/corr/abs-2402-19446,DBLP:journals/corr/abs-2312-14878,DBLP:journals/corr/abs-2403-02502,DBLP:journals/corr/abs-2311-18232}.
Specifically, they involve training LLMs or agents through exploration and learning. 
As a representative method, RL has succeeded in LLM alignment \citep{DBLP:journals/corr/abs-2112-00861,DBLP:journals/corr/abs-2204-05862,DBLP:conf/nips/Ouyang0JAWMZASR22,DBLP:journals/corr/abs-2307-04964,DBLP:journals/corr/abs-2401-06080}, and has been introduced to reasoning and agent tasks, achieving excellent results \citep{DBLP:journals/corr/abs-2402-05808,luong2024reft,DBLP:journals/corr/abs-2402-19446,DBLP:journals/corr/abs-2312-14878}. However, in our multi-environment scenarios, reward consistency and training stability can become problematic \citep{DBLP:journals/corr/abs-2402-19446,DBLP:journals/corr/abs-2403-02502,DBLP:journals/corr/abs-2401-07382}. 
Another line of work uses self-evolution/self-improvement, where the model explores the environment to obtain high-reward trajectories and fine-tunes itself based on these trajectories, achieving promising performance in reasoning, coding, and web tasks \citep{DBLP:journals/corr/abs-2308-08998,DBLP:journals/corr/abs-2312-06585,DBLP:conf/nips/ZelikmanWMG22,DBLP:journals/corr/abs-2308-01825,DBLP:journals/corr/abs-2312-10003,DBLP:journals/corr/abs-2403-14589,DBLP:journals/corr/abs-2403-02502,tao2024survey,tian2024toward,DBLP:journals/corr/abs-2404-03648}. 
However, like RL-based methods, these works only explore within a single environment or task. 
With \AgentGym, this work explores agent evolution using the novel \AgentEvol method, conducting exploration across multiple environments. 

\section{Conclusion}
In this work, we present a new framework named \AgentGym that includes an interactive platform with diverse agent environments and tasks, a benchmark named \AgentEval, and trajectory sets called \AgentTraj and \AgentTraj-L. Additionally, we propose a novel algorithm \AgentEvol, and take the initial step in exploring the self-evolution of generally-capable LLM-based agents across multiple environments. Empirical results demonstrate the effectiveness of our framework and our method. We also perform sufficient ablation and analysis to investigate how our method works. We hope our work can help the AI community develop more advanced generalist LLM-based agents.

\bibliography{main}

\clearpage
\appendix
\section{Limitations} \label{appendix:limitations}
This paper proposes a novel framework named \AgentGym. It includes an interactive platform with diverse environments and tasks, an agent benchmark \AgentEval, and two collections of expert trajectories \AgentTraj and \AgentTraj-L. Additionally, we introduce a novel algorithm, \AgentEvol, to explore the evolutionary capabilities of generally-capable LLM-based agents. Despite the contributions and the fact that our method performs well, our work still has some limitations. \textbf{Firstly}, for computational efficiency, we do not perform multiple samplings in each iteration. However, in the analysis in Section \ref{sec:Discussion_and_Analysis}, we find that more sampling leads to better results, although the improvement is not significant. In the future, we hope to increase the number of samples $K$ to a larger value when sufficient computational resources are available, to explore the upper limits of our method. \textbf{Secondly}, although we validate the effectiveness of our method on three different models (Llama2-Chat-7B, Llama-2-Chat-13B, and DeepSeek-Coder-1.3B), we hope to verify it on stronger and larger base models in the future to explore the potential for building more generally-capable agents. 

\section{Broader Impacts} \label{appendix:broader impacts}
The proposed \AgentGym and \AgentEvol facilitate the self-evolution of generally-capable agents, and our focus is on the self-evolution of capabilities, but it is crucial to consider safety and ethical issues during usage. Agents must not be allowed to violate human values. Therefore, it is essential to strengthen supervision and regulation when eliciting agents' self-evolution capabilities. In the future, we hope to improve the framework's functionality to align agents with human values.

\section{Details of Environments in \AgentGym} \label{sec: env details}
\paragraph{WebShop (WS) \citep{DBLP:conf/nips/Yao0YN22}.}
WebShop is an interactive web environment for web shopping. The agents are given instructions, and need to buy a product that matches the specifications. The agents can click a button on the webpage or search for something by the search engine. 
WebShop contains $12$k instructions and provides over one million real products from amazon.com. 
We select $6910$ instructions. 
For \AgentTraj, we collect $1000$ trajectories with SOTA models ($700$) and human annotations ($300$). 
For \AgentTraj-L, we collect $3930$ trajectories with SOTA models ($3430$) and human annotations ($500$). 
We take the success rate as the evaluation metric and set the maximum round to $10$.\footnote{https://github.com/princeton-nlp/WebShop/blob/master/LICENSE.md}

\paragraph{WebArena (WA) \citep{DBLP:journals/corr/abs-2307-13854}.}
WebArena is a realistic and reproducible web environment. It contains four categories: E-commerce platforms, social forum platforms, collaborative development platforms, and content management systems. It supports 12 different web browsing actions. The observation space consists of a web page URL, the opened tabs, and the web page content. Completing tasks in this highly realistic environment requires the agent to possess strong memory, high-level planning, common sense, and reasoning abilities. The reward from the environment is consistent with the original paper. We filter 20 evaluating test instances from the original dataset for three main sub-tasks: Information-seeking, Site Navigation and Content \& configuration operation. 
We take the success rate as the evaluation metric and set the maximum round to $25$.\footnote{https://github.com/web-arena-x/webarena/blob/main/LICENSE}

\paragraph{MAZE (MZ) \citep{abdulhai2023lmrl}.}
MAZE is a word game. Agents, acting as players, can know their own position, the goal position, and the directions where there are walls around them. Agents decide to move one square in one of four directions (up, down, left, or right) each time, receiving a reward of -1 for every move until they reach the goal position. We use GPT-4-Turbo to add thoughts to the trajectories sampled by LMRL-Gym and create our dataset. 
For \AgentTraj, we include $100$ trajectories. For \AgentTraj-L, we include $215$ trajectories.
We take the success rate as the evaluation metric and set the maximum round to $15$.\footnote{https://github.com/abdulhaim/LMRL-Gym/blob/main/LICENSE}

\paragraph{Wordle (WD) \citep{abdulhai2023lmrl}.}
Wordle is a word-guessing game that tests agents' ability to reason at the level of individual letters. Agents guess the target word from a given vocabulary containing some five-letter words. After each guess, agents are told whether each letter in the guessed word is in the target word and whether its position is correct and receive a reward of -1 for each step until they guess the target word or run out of attempts. We take the success rate as the evaluation metric and set the maximum round to $8$. We also use GPT-4-Turbo to add thoughts to the trajectories sampled by LMRL-Gym. For \AgentTraj, we include $500$ trajectories. For \AgentTraj-L, we include $955$ trajectories.

\paragraph{ALFWorld (ALF) \citep{DBLP:conf/iclr/ShridharYCBTH21}.}
ALFWorld is a household environment based on TextWorld, where agents need to explore rooms and use common sense reasoning to execute tasks. The action space of ALFWorld includes picking up and placing items, observing surroundings, using furniture, and more. The environment provides feedback on the execution of actions based on predefined logic. We take the success rate as the evaluation metric and set the maximum round to $30$. ALFWorld have six types of tasks. We get $3827$ instructions from the original work. 
For \AgentTraj, we collect $500$ trajectories with SOTA models($400$) and human annotations ($100$).
For \AgentTraj-L, we collect $2420$ trajectories with SOTA models($1920$) and human annotations ($500$).
\footnote{https://github.com/alfworld/alfworld/blob/master/LICENSE}

\paragraph{SciWorld (Sci) \citep{DBLP:conf/emnlp/WangJCA22}.}
ScienceWorld is a benchmark for testing agents' scientific reasoning abilities in a new interactive text environment at the standard elementary science curriculum level. ScienceWorld includes 30 types of tasks, such as using measurement instruments and conducting mechanics experiments. Its action space is task-related, with the environment simulator providing the effects of actions. Because the ScienceWorld repository provides golden paths and existing models cannot achieve high performance, we use GPT-4-Turbo to generate thoughts for golden paths of 22 types of interactions that are not too long. 
For \AgentTraj, we include $1000$ trajectories. For \AgentTraj-L, we include $2120$ trajectories.
We take reward as the evaluation metric and set the maximum round to $30$.\footnote{https://github.com/allenai/ScienceWorld/blob/main/LICENSE}

\paragraph{BabyAI (Baby) \citep{DBLP:conf/iclr/Chevalier-Boisvert19}.}
The BabyAI platform is an interactive grid world simulator with 40 instruction-following tasks where the agent is asked to interact with objects. The agent has a limited 7x7 sight of view and can only operate objects in front. The original implementation of BabyAI presents observations in the form of images and low-level actions like "move forward" and "turn left". The implementation from AgentBoard converts graphic observations into textual instructions and expands the action space with high-level actions like "pickup green key 1" and“go through blue locked door 2". The agent receives a non-zero reward discounted by the number of steps when reaching the goal, and 0 otherwise. 
For \AgentTraj, we annotate $400$ trajectories of 18 out of all 40 tasks with SOTA models. 
For \AgentTraj-L, we annotate $810$ trajectories with SOTA models. 
We take reward as the evaluation metric and set the maximum round to $20$.\footnote{https://github.com/mila-iqia/babyai/blob/master/LICENSE}

\paragraph{TextCraft (TC) \citep{DBLP:journals/corr/abs-2311-05772}.}
Similar to WordCraft, TextCraft is a text-only environment for crafting Minecraft items. This environment constructs a crafting tree based on Minecraft's crafting recipes, comprising 544 nodes, each representing a target item. In TextCraft, each task provides a specific target item alongside a list of crafting commands generated by the tree. These tasks are structured compositionally, incorporating crafting recipes of varying complexity ranging from 1 to 4 steps. The environment supports three valid actions: craft <item> using <ingredients>, get <item>, and inventory. Each round, the environment checks the agent's actions and returns the execution state. Apart from craftable items and their ingredients, all other items are obtainable from the environment. Agents can get a reward of 1 only upon successfully crafting the target item. We select 100 tasks for the test set and use the remaining tasks for training. 
For \AgentTraj, we annotate $300$ trajectories with SOTA models ($254$) and human annotation ($46$), with every action in the trajectories verified by the environment.
For \AgentTraj-L, we annotate $374$ trajectories with SOTA models ($299$) and human annotation ($75$).
We take the success rate as the evaluation metric and set the maximum round to $20$.\footnote{https://github.com/archiki/ADaPT/blob/main/LICENSE}

\paragraph{Weather (WT) \citep{DBLP:journals/corr/abs-2401-13178}.}
The Weather Environment allows LLM agents to utilize a weather tool to access data on temperature, precipitation, and air quality for various locations and time periods. It includes 18 different actions that agents can use to achieve weather-related objectives. This environment leverages Python code to integrate the Open-Meteo API and implement the necessary functions. If the agent's final answer matches the reference answer, it receives a reward of 1; otherwise, it receives a reward of $0$. We expand the original dataset of 20 queries to a total of $343$ queries by using GPT-3.5-Turbo and GPT-4-Turbo for augmentation using self-instruct and instruction evolution. 
Finally, we select 20 questions as the evaluating set, leaving the remaining questions as the training set.
For \AgentTraj, we annotate $160$ trajectories with SOTA models ($140$) and human annotators ($20$). We also refine the annotations with human review to ensure accuracy. 
For \AgentTraj-L, we annotate $311$ trajectories with SOTA models ($230$) and human annotators ($81$). 
We take the success rate as the evaluation metric and set the maximum round to $10$.\footnote{https://github.com/hkust-nlp/AgentBoard. The codebase is licensed under an Apache-2.0 License and the dataset is licensed under a GNU General Public License, version 2.}

\paragraph{Movie (MV) \citep{DBLP:journals/corr/abs-2401-13178}.}
The Movie Environment grants LLM agents to utilize the movie tool for accessing cinematic data, including film details, personnel, and production companies. It offers 16 distinct actions that agents can use to achieve various movie-related objectives. This tool integrates the API and data from The Movie Database, implementing the necessary functions to establish its capabilities. If the agent's final answer matches the reference answer, it receives a reward of 1; otherwise, it receives a reward of 0. To enhance the dataset, we expand the original 20 questions to $238$ by using GPT-3.5-Turbo and GPT-4-Turbo for query augmentation. GPT-4-Turbo is employed to annotate $100$ trajectories in \AgentTraj, and the annotations are further corrected through human annotations to ensure accuracy. 
We also use GPT-4-Turbo to annotate $215$ trajectories for \AgentTraj-L.
We select 20 questions for the evaluating set, with the remaining questions designated as the training set. We take the success rate as the evaluation metric and set the maximum round to $12$.

\paragraph{Academia (AM) \citep{DBLP:journals/corr/abs-2401-13178}.}
The Academia Environment equips LLM agents with the academic tools to query information related to computer science research, including academic papers and author details. It offers 7 different actions for achieving various academic research objectives. During its development, it utilizes data from the Citation Network Dataset, crafts the necessary functions, and subsequently constructs the Academia tool. If the agent's final answer matches the reference answer, it receives a reward of 1; otherwise, it receives a reward of 0. 
% We take this environment as a held-out one, and annotate no trajectory. 
The original 20 questions are used as the evaluating set. We take the success rate as the evaluation metric and set the maximum round to $12$.

\paragraph{TODOList (TL) \citep{DBLP:journals/corr/abs-2401-13178}.}
The TodoEnvironment enables LLM agents to query and amend personal agenda data through the todo tool, offering 11 different actions. This tool is implemented based on the TodoList API. If the agent's final answer or operations matches the reference ones, it receives a reward of 1; otherwise, it receives a reward of 0. To enhance the dataset, we expand the original 20 questions to 155 using GPT-3.5-Turbo and GPT-4-Turbo for data augmentation. 
For \AgentTraj, we annotate $70$ trajectories with GPT-4-Turbo.
For \AgentTraj-L, we annotate the queries to get $135$ trajectories with GPT-4-Turbo ($96$) and human annotators ($39$).
The annotations are further refined by human review to ensure accuracy. Finally, we select 20 questions for the evaluating set, with the remaining questions designated as the training set. We take the success rate as the evaluation metric and set the maximum round to $15$.

\paragraph{Sheet (ST) \citep{DBLP:journals/corr/abs-2401-13178}.}
The Sheet Environment allows LLM agents to use the sheet tool to access and modify spreadsheet data, providing 20 different actions for operating on an Excel sheet. This tool is built upon the Google Sheets API. The reward returned by the environment is based on the similarity between the table manipulated by the agent and the reference table, with a value range of $[0,1]$. 
The original $20$ questions are used as the evaluating set. We take reward as the evaluation metric and set the maximum round to $15$. 

\paragraph{BIRD (BD) \citep{DBLP:conf/nips/ZhengC00WZL0LXZ23}.} 
Code ability is a crucial aspect of capability for LLM-based agents. In this environment, we focus on database management ability. We wrap the BIRD-SQL dataset and provide a unified API for agents to interact with. BIRD-SQL is a bench for large-scale database-grounded text-to-SQL evaluation. It requires the agent to query a database using a SELECT statement to get the correct answer. It contains $9428$ unique problems with a golden answer for training. We select $3200$ of them as the instruction set.
For \AgentTraj, we employ GPT-4-Turbo to add thoughts for $2000$ of the training set problems. 
For \AgentTraj-L, we employ GPT-4-Turbo to add thoughts for $3000$ of the training set problems. 
We take success rate as the evaluation metric and the maximum round is $1$ as BD is a single-round programming task.\footnote{https://github.com/AlibabaResearch/DAMO-ConvAI/tree/main/bird. The bench is under a CC BY-NC 4.0 License.}

\section{Platform Architecture of \AgentGym}\label{appendix:platform_arch}
\begin{figure*}[t]
    \includegraphics[width=0.8\linewidth]{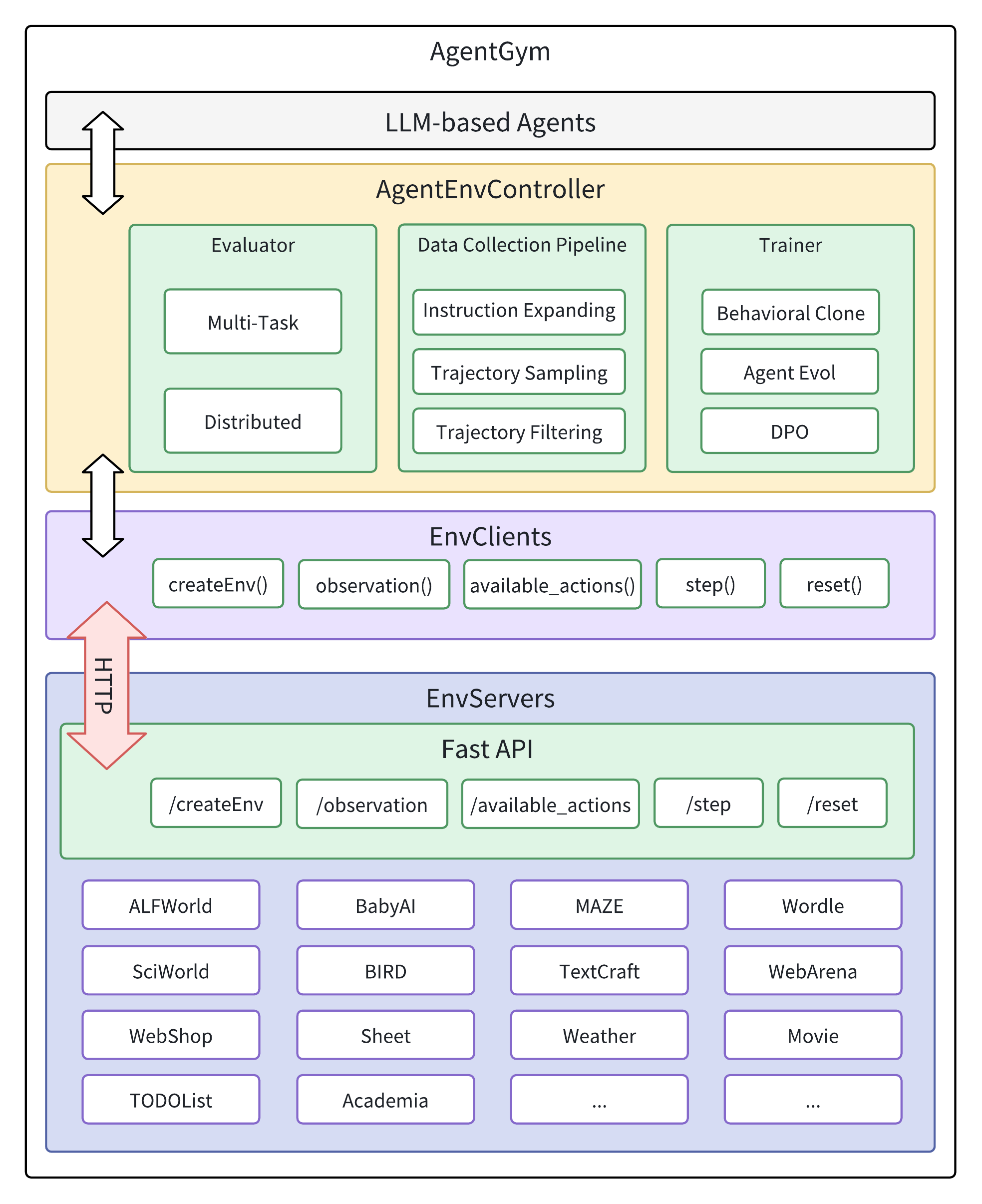}
    \centering
    \vspace{-0.3cm}
 	\caption{
   An illustration of the architecture of \AgentGym platform.
  }\label{fig:platform_arch}
  \vspace{-0.18cm}
\end{figure*} 

The platform architecture of \AgentGym is illustrated in Figure \ref{fig:platform_arch}.In \AgentGym, different environments are deployed on different servers or ports and provide encapsulated HTTP services externally. This decouples the environments from other parts. These services include APIs such as \texttt{/createEnv} to create an environment, \texttt{/observation} to get the current observation from the environment, \texttt{/available\_actions} to get the currently available actions, \texttt{/step} to perform an action, and \texttt{/reset} to reset the environment. We have implemented $14$ types of environments and $89$ tasks, and developers can easily develop new environments and add them to \AgentGym by encapsulating the aforementioned interfaces. EnvClients have the responsibility of receiving services provided by the server and encapsulating them into functions for user calls. AgentController is our core component that connects the agent and the environment. It is responsible for evaluating the agent, collecting data, and training the agent.

\section{More Implementation Details}\label{appendix:more_impl_details}
The learning rate for both the BC phase and the evolution phase is set to \(1 \times 10^{-5}\). 
For the BC baseline on the complete trajectory set, we run $3$ epochs, while for evolution, we run 1 epoch per iteration. 
When evaluating models that have not been fine-tuned on expert trajectories, we use a few-shot approach; when evaluating models that have been trained on expert trajectories, we use a zero-shot approach. In evaluation, we set the temperature to $0$. In the exploration step of \AgentEvol, we set the temperature to $0.7$. 
With eight A100-80GB GPUs, the complete evolution process based on Llama-2-Chat-7B (four iterations) takes approximately twenty hours, and testing takes about one hour.
When performing evolution with both successful and failed trajectories, we include a BC objective to stabilize the training procedure following previous work \citep{DBLP:journals/corr/abs-2404-03648}. The DPO objective and the BC objective are assigned equal weights. We use a learning rate of \(1 \times 10^{-6}\) here because it provides more stable training.

\section{More Experiments}
\subsection{Interactive Rounds in Main Experiments}\label{appendix: Interactive Rounds}
Interactive rounds reflect the efficiency of an agent in solving tasks. Table \ref{table:interactive rounds} shows the interactive rounds of each model/agent across tasks. We also present the evaluation performance in Table \ref{table:interactive rounds} for better and clearer illustration.
We find that agents trained with \AgentTraj-L and \AgentEvol both demonstrate high efficiency, indicating that they can complete tasks in a small number of rounds. Additionally, we observe a trend: agents that require fewer interactive rounds to complete the same task generally perform better. This may be because underperforming agents often struggle to find the optimal path to achieve the final goal or exceed the maximum number of rounds. For example, in ALFWorld and BabyAI, \AgentEvol achieves the best performance as well as the fewest interactive rounds.

\newcommand{\bluenumber}[1]{%
    \textcolor[HTML]{1c7ed6}{#1} 
}

\begin{table*}[t]
\centering
\caption{Evaluating performance and interactive rounds on diverse tasks. The first row of each method indicates performance, while the second row of each method shows the number of \bluenumber{interaction rounds} between the model/agent and the environment.}
\resizebox{0.9999\textwidth}{!}{ 
\begin{tabular}{lccccccccccc}
\toprule
Method &WS &ALF &TC &Sci &Baby &MZ &WD &WT &MV &TL &BD \\

\cmidrule(l){1-1} 
\cmidrule(l){2-12}
\multicolumn{12}{c}{\textbf{Close-sourced Models \& Agents}} \\ 
% Deepseek-chat  & $6.9$ & $20.4$& $15.1$& $20.7$& $11.7$& $14.5$& $5.2$& $6.1$& $5.9$& $4.4$& $1.0$ \\
DeepSeek-Chat & $11.00$ & $51.00$ & $23.00$ & $\mathbf{16.80}$ & $45.67$ & $4.00$ & $24.00$ & $70.00$ & $70.00$ & $75.00$ & $13.50$ \\
& \bluenumber{6.9} & \bluenumber{20.4} & \bluenumber{15.1} & \bluenumber{20.7} & \bluenumber{11.7} & \bluenumber{14.5} & \bluenumber{5.2} & \bluenumber{6.1} & \bluenumber{5.9} & \bluenumber{4.4} & \bluenumber{1.0} \\
Claude-3-Haiku & $5.50$ & $0.00$ & $0.00$ & $0.83$ & $1.93$ & $4.00$ & $16.00$ & $55.00$ & $50.00$ & $65.00$ & $13.50$ \\
& \bluenumber{8.0} & \bluenumber{30.0} & \bluenumber{20.0} & \bluenumber{29.8} & \bluenumber{19.9} & \bluenumber{14.4} & \bluenumber{5.7} & \bluenumber{7.3} & \bluenumber{6.0} & \bluenumber{4.0} & \bluenumber{1.0} \\
Claude-3-Sonnet & $1.50$ & $13.00$ & $38.00$ & $2.78$ & $\mathbf{79.25}$ & $0.00$ & $36.00$ & $65.00$ & $80.00$ & $80.00$ & $\mathbf{17.00}$ \\
& \bluenumber{9.5} & \bluenumber{27.9} & \bluenumber{14.6} & \bluenumber{28.7} & \bluenumber{6.6} & \bluenumber{15.0} & \bluenumber{5.2} & \bluenumber{6.9} & \bluenumber{5.1} & \bluenumber{4.5} & \bluenumber{1.0} \\
GPT-3.5-Turbo & $12.50$ & $26.00$ & $47.00$ & $7.64$ & $71.36$ & $4.00$ & $20.00$ & $25.00$ & $70.00$ & $40.00$ & $12.50$ \\
& \bluenumber{4.9} & \bluenumber{25.2} & \bluenumber{13.1} & \bluenumber{16.5} & \bluenumber{8.4} & \bluenumber{14.4} & \bluenumber{5.3} & \bluenumber{6.6} & \bluenumber{4.6} & \bluenumber{3.4} & \bluenumber{1.0} \\
GPT-4-Turbo & $\mathbf{15.50}$ & $\mathbf{67.50}$ & $\mathbf{77.00}$ & $14.38$ & $72.93$ & $\mathbf{68.00}$ & $\mathbf{88.00}$ & $\mathbf{80.00}$ & $\mathbf{95.00}$ & $\mathbf{95.00}$ & $16.00$ \\
& \bluenumber{8.2} & \bluenumber{18.3} & \bluenumber{9.9} & \bluenumber{18.1} & \bluenumber{9.1} & \bluenumber{9.0} & \bluenumber{4.0} & \bluenumber{6.0} & \bluenumber{4.5} & \bluenumber{4.0} & \bluenumber{1.0} \\
% \midrule
\midrule
% \hdashline
\multicolumn{12}{c}{\textbf{Open-sourced Models \& Agents}} \\ 
Llama2-Chat-7B &$0.50$ & $2.00$& $0.00$& $0.83$& $0.23$& $0.00$& $0.00$& $0.00$& $0.00$& $0.00$& $1.50$\\
& \bluenumber{6.4} & \bluenumber{22.6} & \bluenumber{14.5} & \bluenumber{27.5} & \bluenumber{9.5} & \bluenumber{15.0} & \bluenumber{6.0} & \bluenumber{9.9} & \bluenumber{12.0} & \bluenumber{15.0} & \bluenumber{1.0} \\
Llama2-Chat-13B  &$1.00$ & $3.50$& $0.00$& $0.83$& $0.10$& $0.00$& $0.00$& $0.00$& $0.00$& $0.00$& $1.50$ \\
& \bluenumber{8.1} & \bluenumber{19.6} & \bluenumber{16.5} & \bluenumber{21.3} & \bluenumber{10.9} & \bluenumber{13.4} & \bluenumber{6.0} & \bluenumber{10.0} & \bluenumber{12.0} & \bluenumber{15.0} & \bluenumber{1.0} \\

AgentLM-7B & $36.50$ & $71.00$ & $\mathbf{4.00}$ & $1.63$ & $0.49$ & $\mathbf{12.00}$ & $4.00$ & $0.00$ & $\mathbf{5.00}$ & $15.00$ & $5.00$  \\
& \bluenumber{4.7} & \bluenumber{17.7} & \bluenumber{19.4} & \bluenumber{28.5} & \bluenumber{7.5} & \bluenumber{13.9} & \bluenumber{2.0} & \bluenumber{8.3} & \bluenumber{11.7} & \bluenumber{10.6} & \bluenumber{1.0} \\
AgentLM-13B  & $39.50$ & $\mathbf{73.00}$ & $0.00$ & $2.75$ & $0.45$ & $8.00$ & $0.00$ & $\mathbf{10.00}$ & $\mathbf{5.00}$ & $5.00$ & $3.00$  \\
& \bluenumber{4.8} & \bluenumber{17.8} & \bluenumber{19.4} & \bluenumber{28.5} & \bluenumber{7.6} & \bluenumber{13.9} & \bluenumber{6.0} & \bluenumber{6.6} & \bluenumber{10.7} & \bluenumber{8.4} & \bluenumber{1.0} \\
AgentLM-70B  & $\mathbf{49.50}$ & $67.00$ & $\mathbf{4.00}$ & $\mathbf{10.68}$ & $\mathbf{0.66}$ & $8.00$ & $4.00$ & $0.00$ & $0.00$ & $\mathbf{40.00}$ & $\mathbf{7.50}$  \\
& \bluenumber{4.9} & \bluenumber{18.5} & \bluenumber{18.8} & \bluenumber{28.2} & \bluenumber{6.3} & \bluenumber{13.9} & \bluenumber{5.2} & \bluenumber{6.6} & \bluenumber{11.6} & \bluenumber{6.7} & \bluenumber{1.0} \\
% \midrule
\midrule
% \hdashline
\multicolumn{12}{c}{\textbf{Ours}} \\ 
% BC$_\text{\ full}$
BC$_{base}$ &$66.50$ & $77.50$& $44.00$& $26.42$& $69.31$& $\mathbf{12.00}$& $12.00$& $25.00$& $5.00$& $45.00$& $8.00$ \\
& \bluenumber{5.6} & \bluenumber{16.4} & \bluenumber{13.7} & \bluenumber{21.3} & \bluenumber{6.7} & \bluenumber{14.3} & \bluenumber{5.9} & \bluenumber{6.2} & \bluenumber{10.8} & \bluenumber{5.4} & \bluenumber{1.0} \\
BC$_{large}$ & $73.50$ & $83.00$ & $60.00$ & $\mathbf{74.47}$ & $74.19$ & $\mathbf{12.00}$ & $\mathbf{36.00}$ & $\mathbf{45.00}$ & $5.00$ & $65.00$ & $8.50$  \\
& \bluenumber{5.5} & \bluenumber{16.1} & \bluenumber{14.3} & \bluenumber{29.3} & \bluenumber{6.2} & \bluenumber{14.3} & \bluenumber{5.7} & \bluenumber{6.4} & \bluenumber{10.2} & \bluenumber{5.0} & \bluenumber{1.0} \\
\textbf{\AgentEvol} &$\mathbf{76.50}$ & $\mathbf{88.00}$& $\mathbf{64.00}$& $38.00$& $\mathbf{82.70}$& $\mathbf{12.00}$ & $12.00$& $25.00$& $\mathbf{60.00}$& $\mathbf{70.00}$& $\mathbf{9.00}$ \\
& \bluenumber{5.1} & \bluenumber{14.0} & \bluenumber{11.8} & \bluenumber{18.9} & \bluenumber{4.3} & \bluenumber{13.8} & \bluenumber{5.7} & \bluenumber{5.9} & \bluenumber{3.2} & \bluenumber{5.1} & \bluenumber{1.0} \\

\bottomrule
\end{tabular}
}
\label{table:interactive rounds}
\end{table*}

\subsection{Case Study}\label{Appendix:case study}
\begin{figure*}[t]
    \includegraphics[width=0.9\linewidth]{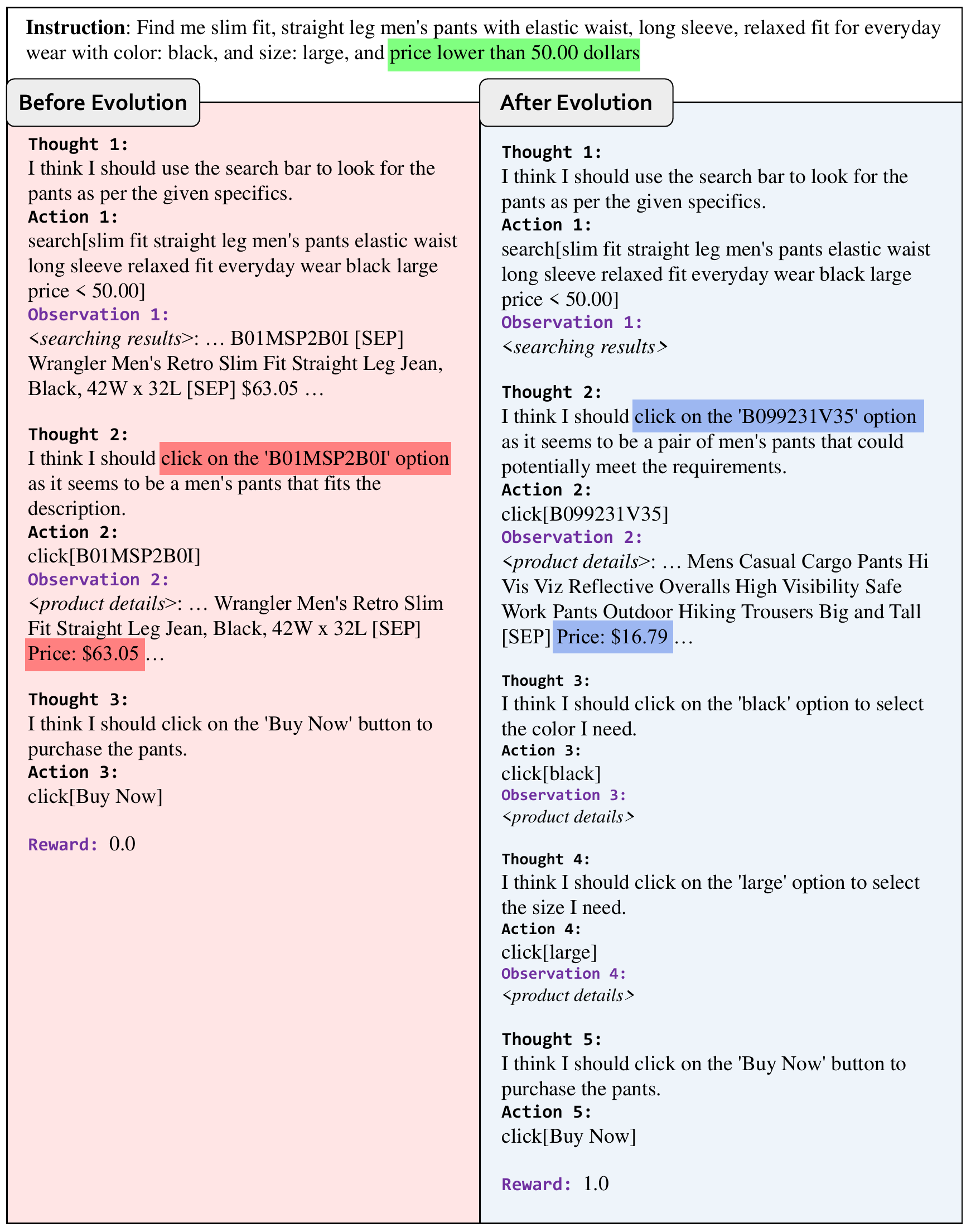}
    \centering
    \vspace{-0.3cm}
 	\caption{
   Case study of WebShop. Before evolution, the agent can not effectively utilize specific information from the environment's feedback and directly chooses an item that exceeds the target price, resulting in task failure. After the evolution, the agent is able to engage in multiple rounds of interaction with the environment, accurately parse the details of the items returned by the environment, and select a product with the correct color, size, and price attributes.
  }\label{fig:case_study_1}
  \vspace{-0.18cm}
\end{figure*} 

\begin{figure*}[t]
    \includegraphics[width=0.9\linewidth]{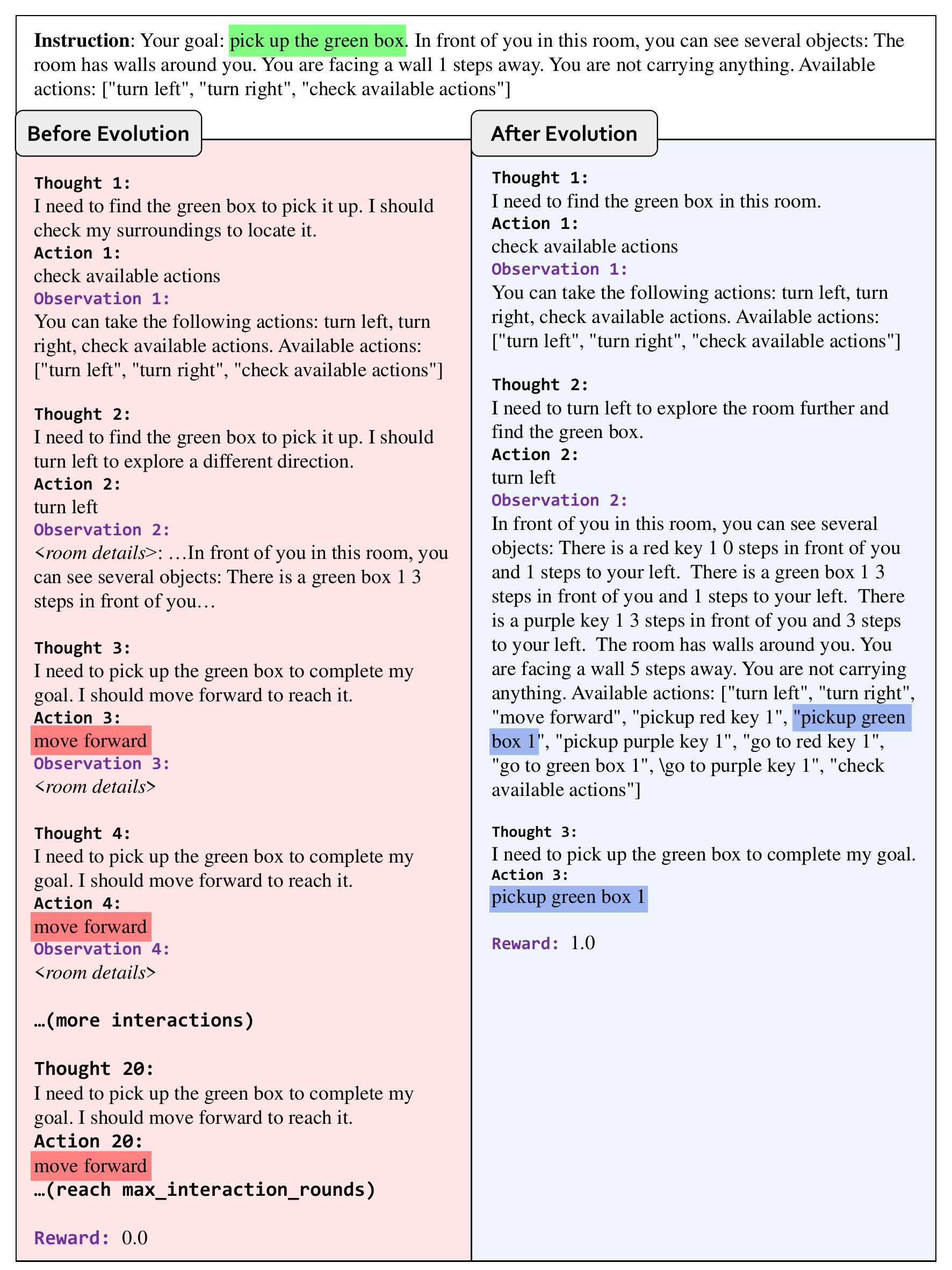}
    \centering
    \vspace{-0.3cm}
 	\caption{
   Case study of BabyAI. The agent before evolution cannot effectively understand spatial relationships and fails to perceive that the target object is right in front of it, leading to incorrect decisions. After receiving the positional information returned by the environment, it repeatedly moves forward until it reaches the interaction limit. After evolution, the agent can accurately determine its position and directly execute the correct "pickup green box 1" action.
  }\label{fig:case_study_2}
  \vspace{-0.18cm}
\end{figure*}

Here, we select two cases to demonstrate the performance comparison before and after the agent evolution, illustrating the effectiveness of \AgentEvol. 

The first case is shown in Figure \ref{fig:case_study_1}. In this case, the user's instruction is ``Find me slim fit, straight leg men's pants with elastic waist, long sleeve, relaxed fit for everyday wear with color: black, and size: large, and price lower than $50.00$ dollars.'' 
Before evolution, the agent can not effectively utilize specific information from the environment's feedback and directly chooses an item that exceeds the target price, resulting in task failure.
However, after evolution, the agent is able to engage in multiple rounds of interaction with the environment, accurately parse the details of the items returned by the environment, and select a product with the correct color, size, and price attributes.

The second case comes from the BabyAI environment, as shown in Figure \ref{fig:case_study_2}. In this environment, the agent's task is to pick up the green box in a room. The agent before evolution cannot effectively understand spatial relationships and fails to perceive that the target object is right in front of it, leading to incorrect decisions. After receiving the positional information returned by the environment, it repeatedly moves forward until it reaches the interaction limit. After evolution, the agent can accurately determine its position and directly execute the correct "pickup green box 1" action.

\section{Prompt Details}\label{appendix: Prompt Details}
The prompt for each task comprises two components: the system prompt and the instruction.
The system prompt provides the initial scenario for each task.
The instruction provides specific queries for each task. 
For consistency, the same prompt template is utilized for human annotation, AI-based annotation of trajectories, and evaluation across all tasks.
The prompt details for the WebShop are shown in Table \ref{table:Prompt details for WebShop.}.
Table \ref{table:Prompt details for ALFWorld.} presents the specifications for ALFWorld.
The TextCraft's prompt details are outlined in Table \ref{table:Prompt details for TextCraft.}.
The prompt details for the SciWorld are shown in Table \ref{table:Prompt details for SciWorld.}.
The prompt details for the BabyAI are shown in Table \ref{table:Prompt details for BabyAI.}.
The prompt details for the MAZE are shown in Table \ref{table:Prompt details for MAZE.}.
The prompt details for the Wordle are shown in Table \ref{table:Prompt details for Wordle.}.
The prompt details for the BIRD are shown in Table \ref{table:Prompt details for BIRD.}.
Table \ref{table:Prompt details for WebArena (Part 1/2).} and \ref{table:Prompt details for WebArena (Part 2/2).} show the prompt details for WebArena.
The prompt details for the Weather are shown in Table \ref{table:Prompt details for Weather (Part 1/4).}, \ref{table:Prompt details for Weather (Part 2/4).}, \ref{table:Prompt details for Weather (Part 3/4).}, \ref{table:Prompt details for Weather (Part 4/4).}.
The prompt details for the TODOList are shown in Table \ref{table:Prompt details for TODOList (Part 1/2).}, \ref{table:Prompt details for TODOList (Part 2/2).}.
The prompt details for the Movie are shown in Table \ref{table:Prompt details for Movie (Part 1/3).}, 
\ref{table:Prompt details for Movie (Part 2/3).}, \ref{table:Prompt details for Movie (Part 3/3).}.
The prompt details for the Academia are shown in Table \ref{table:Prompt details for Academia (Part 1/2).}, 
\ref{table:Prompt details for Academia (Part 2/2).}.
The prompt details for the Sheet are shown in Table \ref{table:Prompt details for Sheet (Part 1/4).},
\ref{table:Prompt details for Sheet (Part 2/4).},
\ref{table:Prompt details for Sheet (Part 3/4).},
\ref{table:Prompt details for Sheet (Part 4/4).}.

\begin{table*}[t]
\caption{Prompt details for WebShop.}
    \label{table:Prompt details for WebShop.}
    \centering
    \vspace{2.8mm}
    \begin{tabular}{p{0.96\linewidth}}
\toprule
\textbf{System Prompt:} \\
You are web shopping. I will give you instructions about what to do. You have to follow the instructions. Every round I will give you an observation and a list of available actions, you have to respond to an action based on the state and instruction. You can use search action if the search is available. You can click one of the buttons in clickables. An action should be of the following structure: search[keywords] click[value].
If the action is not valid, perform nothing. Keywords in search are up to you, but the value in click must be a value in the list of available actions. Remember that your keywords in search should be carefully designed. Your response should use the following format: \\ \\
Thought: I think ... \\
Action: click[something]\\ \\

\textbf{Instruction:} \\
WebShop [SEP] Instruction: [SEP] Find me machine wash women's swimsuits \& cover-ups with drawstring closure, elastic waistband, tummy control with color: black, and size: medium, and price lower than 30.00 dollars [SEP] Search \\
\bottomrule
    \end{tabular}
\end{table*}

\begin{table*}[t]
    \centering
    \caption{Prompt details for ALFWorld.}
    \label{table:Prompt details for ALFWorld.}
    \vspace{2.8mm}
    \begin{tabular}{p{0.96\linewidth}}
\toprule
\textbf{System Prompt:} \\ 
Interact with a household to solve a task. Imagine you are an intelligent agent in a household environment and your target is to perform actions to complete the task goal. At the beginning of your interactions, you will be given a detailed description of the current environment and your goal to accomplish. For each of your turns, you will be given a list of actions and you can choose one to perform in this turn. You should choose from two actions: ``THOUGHT'' or ``ACTION''. If you choose ``THOUGHT'', you should first think about the current condition and plan for your future actions, and then output your action in this turn. Your output must strictly follow this format: \\ \\
Thought: your thoughts. \\
Action: your next action.\\ \\
If you choose ``ACTION'', you should directly output the action in this turn. Your output must strictly follow this format: ``Action: your next action''. After each turn, the environment will give you immediate feedback based on which you plan your next few steps. If the environment outputs ``Nothing happened'', that means the previous action is invalid and you should try more options. Reminder: the action must be chosen from the given available actions. Any actions except provided available actions will be regarded as illegal. Think when necessary, try to act directly more in the process. \\ \\
\textbf{Instruction:} \\
You are in the middle of a room. Looking quickly around you, you see a armchair 1, a coffeetable 1, a diningtable 2, a diningtable 1, a drawer 6, a drawer 5, a drawer 4, a drawer 3, a drawer 2, a drawer 1, a dresser 1, a garbagecan 1, a sidetable 1, a sofa 2, a sofa 1, and a tvstand 1. \\ \\
Your task is to: find two tissuebox and put them in coffeetable. \\ \\
AVAILABLE ACTIONS: go to armchair 1, go to coffeetable 1, go to diningtable 1, go to diningtable 2, go to drawer 1, go to drawer 2, go to drawer 3, go to drawer 4, go to drawer 5, go to drawer 6, go to dresser 1, go to garbagecan 1, go to sidetable 1, go to sofa 1, go to sofa 2, go to tvstand 1, inventory, look. \\
\bottomrule
    \end{tabular}
\end{table*}

\begin{table*}[t]
\centering
\caption{Prompt details for TextCraft.}
\label{table:Prompt details for TextCraft.}
\vspace{2.8mm}
\begin{tabular}{p{0.96\linewidth}}
\toprule
\textbf{System Prompt:} \\
You are given a few useful crafting recipes to craft items in Minecraft. Crafting commands are of the format ``craft [target object] using [input ingredients]''. Every round I will give you an observation, you have to respond to an action based on the state and instruction. You can ``get'' an object (ingredients) from the inventory or the environment, look up the game ``inventory'' by inventory, or ``craft'' (target) using any of the crafting commands. You can use ONLY these crafting commands provided, do not use your own crafting commands. However, if the crafting command uses a generic ingredient like ``planks'', you can use special types of the same ingredient e.g. dark oak ``planks'' in the command instead. Your response should use the following format: \\ \\
Thought: ... \\
Action: ... \\ \\

\textbf{Instruction:} \\
Crafting commands:\\
craft 1 golden shovel using 2 stick, 1 gold ingot\\
craft 1 golden chestplate using 8 gold ingot\\
craft 1 golden sword using 1 stick, 2 gold ingot\\
craft 1 netherite ingot using 4 netherite scrap, 4 gold ingot\\
craft 1 light weighted pressure plate using 2 gold ingot\\
craft 1 golden boots using 4 gold ingot\\
craft 1 golden axe using 2 stick, 3 gold ingot\\
craft 9 gold nugget using 1 gold ingot\\
Goal: craft gold nugget. \\
\bottomrule
\end{tabular}
\end{table*}

\begin{table*}[t]
\centering
\caption{Prompt details for SciWorld.}
\label{table:Prompt details for SciWorld.}
\vspace{2.8mm}
\begin{tabular}{p{0.96\linewidth}}
\toprule
\textbf{System Prompt:} \\
You are an agent for the science world. Every round I will give you an observation, you have to respond with an action based on the observation to finish the given task. \\
Here are the actions you may take: \\
\{``action'': ``open/close OBJ'', ``description'': ``open/close a container'',\} \\ 
\{``action'': ``de/activate OBJ'', ``description'': ``activate/deactivate a device'',\} \\ 
\{``action'': ``connect OBJ to OBJ'', ``description'': ``connect electrical components'', \} \\ 
\{``action'': ``disconnect OBJ'', ``description'': ``disconnect electrical components'',\} \\ 
\{``action'': ``use OBJ [on OBJ]'', ``description'': ``use a device/item'',\} \\  
\{``action'': ``look around'', ``description'': ``describe the current room'',\} \\ 
\{``action'': ``look at OBJ'', ``description'': ``describe an object in detail'',\} \\  
\{``action'': ``look in OBJ'', ``description'': ``describe a container's contents'',\} \\ 
\{``action'': ``read OBJ'', ``description'': ``read a note or book'',\} \\ 
\{``action'': ``move OBJ to OBJ'', ``description'': ``move an object to a container'', \} \\ 
\{``action'': ``pick up OBJ'', ``description'': ``move an object to the inventory'', \} \\ 
\{``action'': ``put down OBJ'', ``description'': ``drop an inventory item'',\} \\ 
\{``action'': ``pour OBJ into OBJ'', ``description'': ``pour a liquid into a container'', \} \\ 
\{``action'': ``dunk OBJ into OBJ'', ``description'': ``dunk a container into a liquid'', \} \\ 
\{``action'': ``mix OBJ'', ``description'': ``chemically mix a container'',\} \\ 
\{``action'': ``go to LOC'', ``description'': ``move to a new location'',\} \\ 
\{``action'': ``eat OBJ'', ``description'': ``eat a food'',\} \\ 
\{``action'': ``flush OBJ'', ``description'': ``flush a toilet'',\} \\ 
\{``action'': ``focus on OBJ'', ``description'': ``signal intent on a task object'',\} \\ 
\{``action'': ``wait'', ``description'': ``take no action for 10 iterations'',\} \\ 
\{``action'': ``wait1'', ``description'': ``take no action for 1 iteration'', \} \\ 
\{``action'': ``task'', ``description'': ``describe current task'',\} \\ 
\{``action'': ``inventory'', ``description'': ``list your inventory''\} \\ 
\\
Your response should use the following format:\\ \\
Thought: your thoughts. \\ 
Action: your next action. \\ \\

\textbf{Instruction:} \\
Your task is to find a(n) non-living thing. First, focus on the thing. Then, move it to the orange box in the living room.\
This room is called the bedroom. In it, you see: the agent, a substance called air, a bed. On the bed is: a mattress. On the mattress is: a white pillow.
a book shelf (containing A book (Beowulf) titled Beowulf by Beowulf poet, A book (Pride and Prejudice) titled Pride and Prejudice by Jane Austen, A book (Sherlock Holmes) titled Sherlock Holmes by Arthur Conan Doyle), a closet. The closet door is closed. a finger painting, a table. On the table is: nothing.\
You also see: A door to the hallway (that is closed) \\
\bottomrule
\end{tabular}
\end{table*}

\begin{table*}[t]
\centering
\caption{Prompt details for BabyAI.}
\label{table:Prompt details for BabyAI.}
\vspace{2.8mm}
\begin{tabular}{p{0.96\linewidth}}
\toprule
\textbf{System Prompt:} \\
You are an exploration master who wants to finish every goal you are given. Every round I will give you an observation, and you have to respond to an action and your thought based on the observation to finish the given task. You are placed in a room and you need to accomplish the given goal with actions. You can use the following actions: \

- turn right \\
- turn left \\
- move forward \\
- go to <obj> <id> \\
- pick up <obj> <id> \\
- go through <door> <id>: <door> must be an open door. \\
- toggle and go through <door> <id>: <door> can be a closed door or a locked door. If you want to open a locked door, you need to carry a key that is of the same color as the locked door. \\
- toggle: there is a closed or locked door right in front of you and you can toggle it. \\ \\
Your response should use the following format:\\ \\
Thought: <Your Thought> \\
Action: <Your Action>\\ \\

\textbf{Instruction:} \\
Your goal: go to the red ball \\
In front of you in this room, you can see several objects: There is a grey box 1 1 steps in front of you and 1 steps to your left. There is a grey ball 1 1 steps in front of you and 2 steps to your right. There is a grey key 1 1 steps in front of you and 3 steps to your right. The room has walls around you. You are facing a wall 3 steps away. You are not carrying anything.\\
Available actions: [``turn left'', ``turn right'', ``move forward'', ``pickup red ball 1'', ``pickup red box 1'', ``go to red ball 1'', ``go to red box 1'', ``check available actions''] \\
\bottomrule
\end{tabular}
\end{table*}

\begin{table*}[t]
\centering
\caption{Prompt details for MAZE.}
\label{table:Prompt details for MAZE.}
\vspace{2.8mm}
\begin{tabular}{p{0.96\linewidth}}
\toprule
\textbf{System Prompt:} \\
You are an expert maze solver. Your objective is to reach the goal in as few steps as possible. At each step you will be given information about where the goal is, your current position, and the walls that surround you. When you move right you increase your y position by 1, when you move down you increase your x position by 1. Your possible actions are ``move up'', ``move down'', ``move left'', ``move right''. Formally, your return should be in this format: \\ \\
Thought: <Your Thought> \\
Action: <Your Action>\\ \\
\textbf{Instruction:} \\
Now let's start a new game. Return your action and your thought in the format above strictly. Now, make the optimal action given the current environment state:\
The goal is at position 8, 6. Your current position is at position 1, 1. There are walls to your left, above you, below you.\\
\bottomrule
\end{tabular}
\end{table*}

\begin{table*}[t]
\centering
\caption{Prompt details for Wordle.}
\label{table:Prompt details for Wordle.}
\vspace{2.8mm}
\begin{tabular}{p{0.96\linewidth}}
\toprule
\textbf{System Prompt:} \\
You are an expert wordle player. Welcome to the game of Wordle. Your objective is to guess a hidden 5 letter word. You have 6 attempts to guess it correctly and you should try to guess it in as few attempts as possible. When guessing the word, you should format your word as a space separated sequence of letters, like ``s h i r e'' for example. After guessing the word, you will receive feedback from the game environment in the form of a sequence of 5 space separated letters like ``b y g g b'', where each letter indicates some information about the hidden word. The environment will return one of three letters - ``b'', ``g'', or ``y'' – for each letter in the word you guessed. We describe the meaning of each letter below:

``b'': If the environment returns a ``b'', it means that the letter at that position in your guessed word is not in the hidden word.\\
``y'': If the environment returns a ``y'', it means that the letter at that position in your guessed word is in the hidden word but is not in the correct position.\\
``g'': If the environment returns a ``g'', it means that the letter at that position in your guessed word is in the hidden word and is in the correct position.\\

As a note, if you guess an invalid word (e.g. not a 5 letter word or a word not in the vocabulary), the environment will respond with an ``invalid word'' message. In general though, you should use this information returned by the environment to update your belief about what the hidden word might be and adjust your next guess accordingly.\\ \\
\textbf{Instruction:} \\
Now let's start a new game. Remember, the word you guess should be strictly in the vocabulary. You should return your thought and your word strictly in the formation mentioned above.\\
\bottomrule
\end{tabular}
\end{table*}

\begin{table*}[t]
\centering
\caption{Prompt details for BIRD.}
\label{table:Prompt details for BIRD.}
\vspace{2.8mm}
\begin{tabular}{p{0.96\linewidth}}
\toprule
\textbf{System Prompt:} \\
Given you a description of a SQLite database system, I will ask you a question, then you should help me operate the SQLite database with SQL to answer the question.\\

You have to explain the problem and your solution to me and write down your thoughts. After thinking and explaining thoroughly, you should give a SQL statement to solve the question. Your response should be like this:\\ \\
Thought: Your thought here.\\

Action: \verb|SELECT * FROM table WHERE condition;|\\ \\

You MUST put SQL in markdown format without any other comments. Your SQL should be in one line. Every time you can only execute one SQL statement.\\ \\

\textbf{Instruction:} \\
debit\_card\_specializing contains tables such as customers, gasstations, products, transactions\_1k, yearmonth. Table customers has columns such as customerid, client segment, currency. customerid is the primary key. Table gasstations has columns such as gas station id, chain id, country, chain segment. gas station id is the primary key. Table products has columns such as product id, description. product id is the primary key. Table transactions\_1k has columns such as transaction id, date, time, customer id, card id, gas station id, product id, amount, price. transaction id is the primary key. Table yearmonth has columns such as customer id, date, consumption.  is the primary key. The date of yearmonth is the foreign key of client segment of customers. \\ \\

Among the transactions made in the gas stations in the Czech Republic, how many of them are taken place after 2012/1/1?
\\
\bottomrule
\end{tabular}
\end{table*}

\begin{table*}[t]
\centering
\caption{Prompt details for WebArena (Part 1/2).}
\label{table:Prompt details for WebArena (Part 1/2).}
\vspace{2.8mm}
\begin{tabular}{p{0.96\linewidth}}
\toprule
\textbf{System Prompt:} \\
You are an autonomous intelligent agent tasked with navigating a web browser. You will be given web-based tasks. These tasks will be accomplished through the use of specific actions you can issue.\\ \\
Here's the information you'll have:\\
The user's objective: This is the task you're trying to complete.\\
The current web page's accessibility tree: This is a simplified representation of the webpage, providing key information.\\
The current web page's URL: This is the page you're currently navigating.\\
The open tabs: These are the tabs you have open.\\
The previous action: This is the action you just performed. It may be helpful to track your progress.\\ \\

The actions you can perform fall into several categories:\\ \\

Page Operation Actions:\\
click [id]: This action clicks on an element with a specific id on the webpage.\\
type [id] [content] [press\_enter\_after=0|1]: Use this to type the content into the field with id. By default, the ``Enter'' key is pressed after typing unless press\_enter\_after is set to 0.\\
hover [id]: Hover over an element with id.\\
press [key\_comb]:  Simulates the pressing of a key combination on the keyboard (e.g., Ctrl+v).\\
scroll [direction=down|up]: Scroll the page up or down.\\ \\

Tab Management Actions:\\
new\_tab: Open a new, empty browser tab.\\
tab\_focus [tab\_index]: Switch the browser's focus to a specific tab using its index.\\
close\_tab: Close the currently active tab.\\ \\

URL Navigation Actions:\\
goto [url]: Navigate to a specific URL.\\
go\_back: Navigate to the previously viewed page.\\
go\_forward: Navigate to the next page (if a previous 'go\_back' action was performed).\\ \\

Completion Action:\\
stop [answer]: Issue this action when you believe the task is complete. If the objective is to find a text-based answer, provide the answer in the bracket. If you believe the task is impossible to complete, provide the answer as ``N/A'' in the bracket.\\ \\

Homepage:\\
If you want to visit other websites, check out the homepage at http://homepage.com. It has a list of websites you can visit.\\
http://homepage.com/password.html lists all the account name and password for the websites. You can use them to log in to the websites.\\ \\

To be successful, it is very important to follow the following rules:\\
1. You should only issue an action that is valid given the current observation.\\
2. You should only issue one action at a time.\\
3. You should follow the examples to reason step by step and then issue the next action.\\
4. Generate the action in the correct format. Start with a ``In summary, the next action I will perform is'' phrase, followed by action inside. For example, ``In summary, the next action I will perform is click [1234]''.\\
5. Issue stop action when you think you have achieved the objective. Don't generate anything after stop.\\
\bottomrule
\end{tabular}
\end{table*}

\begin{table*}[t]
\centering
\caption{Prompt details for WebArena (Part 2/2).}
\label{table:Prompt details for WebArena (Part 2/2).}
\vspace{2.8mm}
\begin{tabular}{p{0.96\linewidth}}
\toprule
\textbf{Instruction:} \\
\textnormal{Observation:}\\
\textnormal{Tab 0 (current): Projects · Dashboard · GitLab}\\ \\
\textnormal{[1] RootWebArea 'Projects · Dashboard · GitLab' focused: True}\\
\textnormal{\hspace*{5mm}[271] link 'Skip to content'}\\
\textnormal{\hspace*{5mm}[398] link 'Dashboard'}\\
\textnormal{\hspace*{5mm}[482] button '' hasPopup: menu expanded: False}\\
\textnormal{\hspace*{5mm}[1947] textbox 'Search GitLab' required: False}\\
\textnormal{\hspace*{5mm}[1907] generic 'Use the shortcut key <kbd>/</kbd> to start a search'}\\
... \\

\textnormal{URL: http://gitlab.com/}\\
\textnormal{OBJECTIVE: Checkout merge requests assigned to me}\\
\textnormal{PREVIOUS ACTION: None}\\
% \textnormal{Action: @type [1585] [white desk] [press\_enter\_after=1]@}\\
% \textnormal{Observation:}\\
% \textnormal{WINDOWED PAGE:\{}\\
% \textnormal{Tab 0 (current): Search results for: `white desk'}\\
% \textnormal{[2635] RootWebArea ``Search results for: `white desk'" focused: True}\\
% \textnormal{\hspace*{5mm}[3869] link `My Wish List'}\\
% \textnormal{\hspace*{5mm}[4827] StaticText `Search'}\\
% \textnormal{\hspace*{5mm}[4072] combobox `Search' autocomplete: both hasPopup: listbox required: False expanded: False}\\
% \textnormal{\hspace*{5mm}\hspace*{5mm}[5027] StaticText `white desk'}\\
% \textnormal{\hspace*{5mm}[4830] link `Advanced Search'}\\
% \textnormal{\hspace*{5mm}[4075] button `Search' disabled: True}\\
% \textnormal{\hspace*{5mm}[3729] main `'}\\
% \textnormal{\hspace*{5mm}\hspace*{5mm}[3842] heading ``Search results for: `white desk'"}\\
% \textnormal{\hspace*{5mm}\hspace*{5mm}[3907] StaticText `Items 1-12 of 38823'}\\
% \textnormal{\hspace*{5mm}\hspace*{5mm}[4781] link `Image'}\\
% \textnormal{\hspace*{5mm}\hspace*{5mm}\hspace*{5mm}[4833] img `Image'}\\
% \textnormal{\hspace*{5mm}\hspace*{5mm}[4783] link `Image'}\\
% \textnormal{\hspace*{5mm}\hspace*{5mm}\hspace*{5mm}[4849] img `Image'}\\
% \textnormal{\}}\\
% \textnormal{URL: http://onestopmarket.com/catalogsearch/result/?q=white+desk}\\
% ...
\bottomrule
\end{tabular}
\end{table*}

\begin{table*}[t]
\centering
\caption{Prompt details for Weather (Part 1/4).}
\label{table:Prompt details for Weather (Part 1/4).}
\vspace{2.8mm}
\begin{tabular}{p{0.96\linewidth}}
\toprule
\textbf{System Prompt:} \\
You are an autonomous intelligent agent. You can use actions to help people solve problems. We detail name, description, input(parameters) and output(returns) of each action as follows:\\
Name: get\_user\_current\_date()\\
Description: Get the user's current date.\\
Returns:\\
The current date in `YYYY-MM-DD' format.\\
\\
Name: get\_user\_current\_location()\\
Description: Get the user's current city.\\
Returns:\\
The user's current city.\\
\\
Name: get\_historical\_temp(latitude, longitude, start\_date, end\_date)\\
Description: Get historical temperature data for a specified location and date range.\\
Parameters:\\
- latitude (Type: number): The latitude of the location.\\
- longitude (Type: number): The longitude of the location.\\
- start\_date (Type: string): The start date of the historical data (YYYY-MM-DD).\\
- end\_date (Type: string): The end date of the historical data (YYYY-MM-DD).\\
Returns:\\
Historical temperature data.\\
\\
Name: get\_historical\_rain(latitude, longitude, start\_date, end\_date)\\
Description: Get historical rainfall data for a specified location and date range.\\
Parameters:\\
- latitude (Type: number): The latitude of the location.\\
- longitude (Type: number): The longitude of the location.\\
- start\_date (Type: string): The start date of the historical data (YYYY-MM-DD).\\
- end\_date (Type: string): The end date of the historical data (YYYY-MM-DD).\\
Returns:\\
Historical rainfall data.\\
\\
Name: get\_historical\_snow(latitude, longitude, start\_date, end\_date)\\
Description: Get historical snowfall data for a specified location and date range.\\
Parameters:\\
- latitude (Type: number): The latitude of the location.\\
- longitude (Type: number): The longitude of the location.\\
- start\_date (Type: string): The start date of the historical data (YYYY-MM-DD).\\
- end\_date (Type: string): The end date of the historical data (YYYY-MM-DD).\\
Returns:\\
Historical snowfall data.\\
\\
Name: get\_snow\_forecast(latitude, longitude, start\_date, end\_date)\\
Description: Get snowfall forecast data for a specified location and date range.\\
Parameters:\\
- latitude (Type: number): The latitude of the location.\\
- longitude (Type: number): The longitude of the location.\\
- start\_date (Type: string): The start date of the forecast (YYYY-MM-DD).\\
- end\_date (Type: string): The end date of the forecast (YYYY-MM-DD).\\
Returns:\\
Snowfall forecast data.\\
\bottomrule
\end{tabular}
\end{table*}

\begin{table*}[t]
\centering
\caption{Prompt details for Weather (Part 2/4).}
\label{table:Prompt details for Weather (Part 2/4).}
\vspace{2.8mm}
\begin{tabular}{p{0.96\linewidth}}
\toprule
Name: get\_current\_snow(latitude, longitude, current\_date)\\
Description: Get current snowfall data for a specified location and date.\\
Parameters:\\
- latitude (Type: number): The latitude of the location.\\
- longitude (Type: number): The longitude of the location.\\
- current\_date (Type: string): The current date to retrieve snowfall data (YYYY-MM-DD).\\
Returns:\\
Current snowfall data.\\
\\
Name: get\_current\_temp(latitude, longitude, current\_date)\\
Description: Get current temperature data for a specified location and date.\\
Parameters:\\
- latitude (Type: number): The latitude of the location.\\
- longitude (Type: number): The longitude of the location.\\
- current\_date (Type: string): The current date to retrieve temperature data (YYYY-MM-DD).\\
Returns:\\
Current temperature data.\\
\\
Name: get\_latitude\_longitude(name)\\
Description: Get latitude and longitude information for a specified location name.\\
Parameters:\\
- name (Type: string): The name of the location. (e.g., city name)\\
Returns:\\
latitude and longitude information for the specified location.\\
\\
Name: get\_elevation(latitude, longitude)\\
Description: Get elevation data for a specified location.\\
Parameters:\\
- latitude (Type: number): The latitude of the location.\\
- longitude (Type: number): The longitude of the location.\\
Returns:\\
Elevation data for the specified location.\\
\\
Name: get\_temp\_forecast(latitude, longitude, start\_date, end\_date)\\
Description: Get temperature forecast data for a specified location and date range.\\
Parameters:\\
- latitude (Type: number): The latitude of the location.\\
- longitude (Type: number): The longitude of the location.\\
- start\_date (Type: string): The start date of the forecast (YYYY-MM-DD).\\
- end\_date (Type: string): The end date of the forecast (YYYY-MM-DD).\\
Returns:\\
Temperature forecast data.\\
\\
Name: get\_rain\_forecast(latitude, longitude, start\_date, end\_date)\\
Description: Get rainfall forecast data for a specified location and date range.\\
Parameters:\\
- latitude (Type: number): The latitude of the location.\\
- longitude (Type: number): The longitude of the location.\\
- start\_date (Type: string): The start date of the forecast (YYYY-MM-DD).\\
- end\_date (Type: string): The end date of the forecast (YYYY-MM-DD).\\
Returns:\\
Rainfall forecast data.\\
\bottomrule
\end{tabular}
\end{table*}

\begin{table*}[t]
\centering
\caption{Prompt details for Weather (Part 3/4).}
\label{table:Prompt details for Weather (Part 3/4).}
\vspace{2.8mm}
\begin{tabular}{p{0.96\linewidth}}
\toprule
Name: get\_current\_rain(latitude, longitude, current\_date)\\
Description: Get current rainfall data for a specified location and date.\\
Parameters:\\
- latitude (Type: number): The latitude of the location.\\
- longitude (Type: number): The longitude of the location.\\
- current\_date (Type: string): The current date to retrieve rainfall data (YYYY-MM-DD).\\
Returns:\\
Current rainfall data.\\
\\
Name: get\_distance(latitude1, longitude1, latitude2, longitude2)\\
Description: Calculate the distance between two sets of latitude and longitude coordinates.\\
Parameters:\\
- latitude1 (Type: number): The latitude of the first location.\\
- longitude1 (Type: number): The longitude of the first location.\\
- latitude2 (Type: number): The latitude of the second location.\\
- longitude2 (Type: number): The longitude of the second location.\\
Returns:\\
The distance between the two sets of coordinates in kilometers.\\
\\
Name: get\_historical\_air\_quality\_index(latitude, longitude, start\_date, end\_date)\\
Description: Get historical air quality index data for a specified location and date range.\\
Parameters:\\
- latitude (Type: number): The latitude of the location.\\
- longitude (Type: number): The longitude of the location.\\
- start\_date (Type: string): The start date of the historical data (YYYY-MM-DD).\\
- end\_date (Type: string): The end date of the historical data (YYYY-MM-DD).\\
Returns:\\
Historical air quality index (PM2.5) data.\\
\\
Name: get\_current\_air\_quality\_index(latitude, longitude, current\_date)\\
Description: Get current air quality index data for a specified location and date.\\
Parameters:\\
- latitude (Type: number): The latitude of the location.\\
- longitude (Type: number): The longitude of the location.\\
- current\_date (Type: string): The current date to retrieve air quality index data (YYYY-MM-DD).\\
Returns:\\
Current air quality index (PM2.5) data.\\
\\
Name: get\_air\_quality\_level(air\_quality\_index)\\
Description: Determine the air quality level based on the air quality index (AQI).\\
Parameters:\\
- air\_quality\_index (Type: number): The air quality index (AQI) value.\\
Returns:\\
The air quality level, which can be `good', `fair', `moderate', `poor', `very poor', or `extremely poor'.\\
\\
Name: check\_valid\_actions()\\
Description: Get supported actions for current tool.\\
Returns:\\
- actions (Type: array): Supported actions for current tool.\\
\\
Name: finish(answer)\\
Description: Return an answer and finish the task\\
Parameters:\\
- answer (Type: [`string', `number', `array']): The answer to be returned \\
\bottomrule
\end{tabular}
\end{table*}

\begin{table*}[t]
\centering
\caption{Prompt details for Weather (Part 4/4).}
\label{table:Prompt details for Weather (Part 4/4).}
\vspace{2.8mm}
\begin{tabular}{p{0.96\linewidth}}
\toprule
If you want to get the latitude and longitude information of a city, you must call ``get\_latitude\_longitude'', do not generate it by yourself which maybe wrong.\\
If you are finished, you will call ``finish'' action\\
Please refer to the format of examples below to solve the requested goal. Your response must be in the format of ``Action: [your action] with Action Input: [your action input]''\\ \\
\textbf{Instruction:} \\
Now new trial starts. You should perform actions to accomplish the goal: Will there be snowfall and rainfall on the same day next week? Tell me Yes or No. Give me one action.\\
\bottomrule
\end{tabular}
\end{table*}

\begin{table*}[t]
\centering
\caption{Prompt details for TODOList (Part 1/2).}
\label{table:Prompt details for TODOList (Part 1/2).}
\vspace{2.8mm}
\begin{tabular}{p{0.96\linewidth}}
\toprule
\textbf{System Prompt:} \\
You are an autonomous intelligent agent. You can use actions to help people solve problems. We detail name, description, input(parameters) and output(returns) of each action as follows:\\
Name: get\_user\_current\_date()\\
Description: Get the user's current date.\\
Returns:\\
The current date in `YYYY-MM-DD' format.\\
\\
Name: get\_user\_current\_location()\\
Description: Get the user's current city.\\
Returns:\\
The user's current city.\\
\\
Name: get\_projects()\\
Description: Get all projects in the TodoList account\\
Returns:\\
- Array of objects with properties:\\
  - id (Type: string)\\
  - name (Type: string)\\
  - order (Type: integer)\\
  - color (Type: string)\\
  - is\_favorite (Type: boolean)\\
\\
Name: update\_project(project\_id, is\_favorite)\\
Description: Update a project\\
Parameters:\\
- project\_id (Type: string)\\
- is\_favorite (Type: string, Enum: [True, False])\\
Returns:\\
Information of the updated project\\
\\
Name: get\_tasks(project\_id)\\
Description: Get all tasks for a given project\\
Parameters:\\
- project\_id (Type: string)\\
Returns:\\
- Array of objects with properties:\\
  - id (Type: string)\\
  - project\_id (Type: string)\\
  - order (Type: integer)\\
  - content (Type: string): Name of the task.\\
  - is\_completed (Type: boolean)\\
  - priority (Type: integer): Task priority from 1 (normal) to 4 (urgent).\\
  - due\_date (Type: string): The due date of the task.\\
\\
Name: get\_task\_description(task\_id)\\
Description: Get the description of a specific task in the TodoList account.\\
Parameters:\\
- task\_id (Type: string)\\
Returns:\\
- id (Type: string): Unique identifier of the task.\\
- content (Type: string): Name of the task.\\
- description (Type: string): Description of the task. Including the Place, Tips, etc.\\
\bottomrule
\end{tabular}
\end{table*}

\begin{table*}[t]
\centering
\caption{Prompt details for TODOList (Part 2/2).}
\label{table:Prompt details for TODOList (Part 2/2).}
\vspace{2.8mm}
\begin{tabular}{p{0.96\linewidth}}
\toprule
Name: get\_task\_duration(task\_id)\\
Description: Get the duration of a specific task in the TodoList account.\\
Parameters:\\
- task\_id (Type: string)\\
Returns:\\
- id (Type: string)\\
- content (Type: string): Name of the task.\\
- duration (Type: string): Duration of the task in the format of `amount(unit)'.\\
\\
Name: complete\_task(task\_id)\\
Description: Mark a task as completed\\
Parameters:\\
- task\_id (Type: string)\\
Returns:\\
information of the completed task\\
\\
Name: update\_task(task\_id, due\_date)\\
Description: Update a task\\
Parameters:\\
- task\_id (Type: string)\\
- due\_date (Type: string)\\
Returns:\\
Information of the updated task\\
\\
Name: delete\_task(task\_id)\\
Description: Delete a specific task from the TodoList account.\\
Parameters:\\
- task\_id (Type: string): Unique identifier of the task to delete.\\
Returns:\\
Information of the deleted task.\\
\\
Name: check\_valid\_actions()\\
Description: Get supported actions for current tool.\\
Returns:\\
Supported actions for current tool.\\
\\
Name: finish(answer)\\
Description: Call this action, when find the answer for the current task or complete essential operations.\\
Parameters:\\
- answer (Type: [`string', `number', `array']): If the task is a question answering task, this is the answer to be returned. If the task is an operation task, the answer in `done'\\
\\
\\
If you are finished, you will call ``finish'' action\\
Please refer to the format of examples below to solve the requested goal. Your response must be in the format of ``Action: [your action] with Action Input: [your action input]''\\ \\

\textbf{Instruction:} \\
Now new trial starts. You should perform actions to accomplish the goal: Could you provide the due date for the task `Tidy up the living room' in the Household Chores project? Please answer in `YYYY-MM-DD' format. Give me one action. \\
\bottomrule
\end{tabular}
\end{table*}

\begin{table*}[t]
\centering
\caption{Prompt details for Movie (Part 1/3).}
\label{table:Prompt details for Movie (Part 1/3).}
\vspace{2.8mm}
\begin{tabular}{p{0.96\linewidth}}
\toprule
\textbf{System Prompt:} \\
You are an autonomous intelligent agent. You can use actions to help people solve problems.
We detail name, description, input(parameters) and output(returns) of each action as follows:\\
Name: get\_search\_movie(movie\_name)\\
Description: Search for a movie by name and return basic details\\
Parameters:\\
- movie\_name (Type: string): The name of the movie to search for.\\
Returns:\\
- id : The ID of the found movie.\\
- overview : The overview description of the movie.\\
- title : The title of the movie. \\ \\

Name: get\_movie\_details(movie\_id)\\
Description: Get detailed information about a movie by ID\\
Parameters:\\
- movie\_id (Type: string): The ID of the movie.\\
Returns:\\
- budget : The budget of the movie.\\
- genres : The genres of the movie.\\
- revenue : The revenue of the movie.\\
- vote\_average : The average vote score of the movie.\\
- release\_date : The release date of the movie. \\ \\

Name: get\_movie\_production\_companies(movie\_id)\\
Description: Get the production companies of a movie by its ID\\
Parameters:\\
- movie\_id (Type: string): The ID of the movie.\\
Returns:\\
- production\_companies : The production companies of the movie. \\ \\

Name: get\_movie\_production\_countries(movie\_id)
Description: Get the production countries of a movie by its ID\\
Parameters:\\
- movie\_id (Type: string): The ID of the movie.\\
Returns:\\
- production\_countries : The production countries of the movie. \\ \\

Name: get\_movie\_cast(movie\_id)\\
Description: Retrieve the list of the top 10 cast members from a movie by its ID.\\
Parameters:\\
- movie\_id (Type: string): The ID of the movie.\\
Returns:\\
- cast : List of the top 10 cast members. \\ \\

Name: get\_movie\_crew(movie\_id)\\
Description: Retrieve the list of crew members (limited to 10) from a movie by its ID. The list primarily includes Director, Producer, and Writer roles.\\
Parameters:\\
- movie\_id (Type: string): The ID of the movie.\\
Returns:\\
- crew : List of the top 10 of crew members \\
\bottomrule
\end{tabular}
\end{table*}

\begin{table*}[t]
\centering
\caption{Prompt details for Movie (Part 2/3).}
\label{table:Prompt details for Movie (Part 2/3).}
\vspace{2.8mm}
\begin{tabular}{p{0.96\linewidth}}
\toprule
Name: get\_movie\_keywords(movie\_id)\\
Description: Get the keywords associated with a movie by ID\\
Parameters:\\
- movie\_id (Type: string): The ID of the movie.\\
Returns:\\
- keywords : The keywords associated with the movie. \\ \\

Name: get\_search\_person(person\_name)\\
Description: Search for a person by name.\\
Parameters:\\
- person\_name (Type: string): The name of the person to search for.\\
Returns:\\
- id : The ID of the found person.\\
- name : The name of the person. \\ \\

Name: get\_person\_details(person\_id)\\
Description: Get detailed information about a person by ID\\
Parameters:\\
- person\_id (Type: string): The ID of the person.\\
Returns:\\
- biography : The biography of the person.\\
- birthday : The birthday of the person.\\
- place\_of\_birth : The place of birth of the person. \\ \\

Name: get\_person\_cast(person\_id)\\
Description: Retrieve the top 10 movie cast roles of a person by their ID\\
Parameters:\\
- person\_id (Type: string): The ID of the person.\\
Returns:\\
- cast : A list of movies where the person has acted, limited to top 10 \\ \\

Name: get\_person\_crew(person\_id)\\
Description: Retrieve the top 10 movie crew roles of a person by their ID\\
Parameters:\\
- person\_id (Type: string): The ID of the person.\\
Returns:\\
- crew : A list of movies where the person has participated as crew, limited to top 10 \\ \\

Name: get\_person\_external\_ids(person\_id)\\
Description: Get the external ids for a person by ID\\
Parameters:\\
- person\_id (Type: string): The ID of the person.\\
Returns:\\
- imdb\_id : The IMDB id of the person.\\
- facebook\_id : The Facebook id of the person.\\
- instagram\_id : The Instagram id of the person.\\
- twitter\_id : The Twitter id of the person. \\ \\

Name: get\_movie\_alternative\_titles(movie\_id)\\
Description: Get the alternative titles for a movie by ID\\
Parameters:\\
- movie\_id (Type: string): The ID of the movie.\\
Returns:\\
- titles : The alternative titles of the movie.\\
- id : The ID of the movie. \\
\bottomrule
\end{tabular}
\end{table*}

\begin{table*}[t]
\centering
\caption{Prompt details for Movie (Part 3/3).}
\label{table:Prompt details for Movie (Part 3/3).}
\vspace{2.8mm}
\begin{tabular}{p{0.96\linewidth}}
\toprule
Name: get\_movie\_translation(movie\_id)\\
Description: Get the description translation for a movie by ID\\
Parameters:\\
- movie\_id (Type: string): The ID of the movie.\\
Returns:\\
- translations : The description translation of the movie.\\
- id : The ID of the movie. \\

Name: check\_valid\_actions()\\
Description: Get supported actions for current tool.\\
Returns:\\
- actions (Type: array): Supported actions for current tool. \\ \\

Name: finish(answer)\\
Description: Return an answer and finish the task\\
Parameters:\\
- answer (Type: [`string', `number', `array']): The answer to be returned \\ \\

If you are finished, you will call ``finish" action\\
Please refer to the format of examples below to solve the requested goal. Your response must be in the format of ``Action: [your action] with Action Input: [your action input]" \\ \\

\textbf{Instruction:} \\
Now new trial starts. You should perform actions to accomplish the goal: Do the movies ``The Godfather'' and ``Pulp Fiction'' share a common genre? Please answer me with Yes or No. Give me one action. \\
\bottomrule
\end{tabular}
\end{table*}

\begin{table*}[t]
\centering
\caption{Prompt details for Academia (Part 1/2).}
\label{table:Prompt details for Academia (Part 1/2).}
\vspace{2.8mm}
\begin{tabular}{p{0.96\linewidth}}
\toprule
\textbf{System Prompt:} \
You are an autonomous intelligent agent. You can use actions to help people solve problems. We detail name, description, input(parameters) and output(returns) of each action as follows:\\
Name: loadPaperNet()\\
Description: Load PaperNet. In this net, nodes are papers and edges are citation relationships between papers.\\
\\
Name: loadAuthorNet()\\
Description: Load AuthorNet. In this net, nodes are authors and edges are collaboration relationships between authors.\\
\\
Name: neighbourCheck(graph, node)\\
Description: List the first-order neighbors connect to the node. In paperNet, neigbours are cited papers of the paper. In authorNet, neigbours are collaborators of the author.\\
Parameters:\\
- graph (Type: string, Enum: [PaperNet, AuthorNet]): The name of the graph to check\\
- node (Type: string): The node for which neighbors will be listed\\
Returns:\\
- neighbors (Type: array)\\
\\
Name: paperNodeCheck(node)\\
Description: Return detailed attribute information of a specified paper in PaperNet\\
Parameters:\\
- node (Type: string): Name of the paper.\\
Returns:\\
- authors : The authors of the paper\\
- year : The puslished year of the paper\\
- venue : The published venue of the paper\\
- n\_citation : The number of citations of the paper\\
- keywords : The keywords of the paper\\
- doc\_type : The document type of the paper\\
\\
Name: authorNodeCheck(node)\\
Description: Return detailed attribute information of a specified author in AuthorNet\\
Parameters:\\
- node (Type: string): name of the author.\\
Returns:\\
- name : The name of the author\\
- org : The organization of the author\\
\\
Name: authorEdgeCheck(node1, node2)\\
Description: Return detailed attribute information of the edge between two specified nodes in a AuthorNet.\\
Parameters:\\
- node1 (Type: string): The first node of the edge\\
- node2 (Type: string): The second node of the edge\\
Returns:\\
- papers : All papers that the two authors have co-authored\\
\bottomrule
\end{tabular}
\end{table*}

\begin{table*}[t]
\centering
\caption{Prompt details for Academia (Part 2/2).}
\label{table:Prompt details for Academia (Part 2/2).}
\vspace{2.8mm}
\begin{tabular}{p{0.96\linewidth}}
\toprule
Name: paperEdgeCheck(node1, node2)\\
Description: Return detailed attribute information of the edge between two specified nodes in a PaperNet.\\
Parameters:\\
- node1 (Type: string): The first node of the edge\\
- node2 (Type: string): The second node of the edge\\
Returns:\\
None\\ \\
Name: check\_valid\_actions()\\
Description: Get supported actions for current tool.\\
Returns:\\
- actions (Type: array): Supported actions for current tool.\\
\\
Name: finish(answer)\\
Description: Return an answer and finish the task\\
Parameters:\\
- answer (Type: [`string', `number', `array']): The answer to be returned\\
\\
\\
If you are finished, you will call ``finish'' action\\
Please refer to the format of examples below to solve the requested goal. Your response must be in the format of ``Action: [your action] with Action Input: [your action input]''\\ \\
\textbf{Instruction:} \\
Now new trial starts. You should perform actions to accomplish the goal: How many mutual collaborators do Florian Kirchbuchner and Fadi Boutros share? Please give me a numerical value as an answer. Give me one action. \\
\bottomrule
\end{tabular}
\end{table*}

\begin{table*}[t]
\centering
\caption{Prompt details for Sheet (Part 1/4).}
\label{table:Prompt details for Sheet (Part 1/4).}
\vspace{2.8mm}
\begin{tabular}{p{0.96\linewidth}}
\toprule
\textbf{System Prompt:} \\
You are an autonomous intelligent agent. You can use actions to help people solve problems. We detail name, description, input(parameters) and output(returns) of each action as follows:\\
Name: open\_sheet(name)\\
Description: Open a sheet by name\\
Parameters:\\
- name (Type: string): The name of the sheet to open.\\
Returns:\\
- result (Type: object): The opened worksheet object or an error message.\\
\\
Name: del\_sheet(name)\\
Description: Deletes the specified sheet.\\
Parameters:\\
- name (Type: string): The name of the sheet to be deleted.\\
Returns:\\
- result (Type: object): Whether the operation was successful.\\
\\
Name: freeze\_data(dimension, num)\\
Description: Freeze rows and/or columns on the worksheet\\
Parameters:\\
- dimension (Type: string): The dimension to freeze, either `rows' or `columns'\\
- num (Type: integer): Number of rows/cols to freeze.\\
Returns:\\
- result (Type: object): Whether the operation was successful.\\
\\
Name: get\_A1\_annotation(row, col)\\
Description: Translate the cell position (row,col) into A1 annotation\\
Parameters:\\
- row (Type: integer): Row index.\\
- col (Type: integer): Column index.\\
Returns:\\
- result (Type: string): The A1 notation of the cell or an error message.\\
\\
Name: insert\_cols(values\_list, col\_idx)\\
Description: Insert columns into sheet at specified column index\\
Parameters:\\
- values\_list (Type: array[array[string]]): A list of lists, each list containing one column's values, which can be expressions\\
- col\_idx (Type: integer): Start column to update. Defaults to 1.\\
Returns:\\
- result (Type: object): The updated worksheet data or an error message.\\
\\
Name: insert\_rows(values\_list, row\_idx)\\
Description: Insert rows into sheet at specified row index\\
Parameters:\\
- values\_list (Type: array[array[string]]): A list of lists, each list containing one row's values, which can be expressions\\
- row\_idx (Type: integer): Start row to update. Defaults to 1.\\
Returns:\\
- result (Type: object): The updated worksheet data or an error message.\\
\bottomrule
\end{tabular}
\end{table*}

\begin{table*}[t]
\centering
\caption{Prompt details for Sheet (Part 2/4).}
\label{table:Prompt details for Sheet (Part 2/4).}
\vspace{2.8mm}
\begin{tabular}{p{0.96\linewidth}}
\toprule
Name: delete\_batch\_data(dimension, index\_list)\\
Description: Delete a batch of data in the sheet\\
Parameters:\\
- dimension (Type: string): The dimension to delete, either `row' or `col'.\\
- index\_list (Type: array[integer]): List of the indexes of rows/cols for deletion.\\
Returns:\\
- result (Type: object): The updated worksheet data or an error message.\\
\\
Name: update\_cell(position, value)\\
Description: Update the value of the cell\\
Parameters:\\
- position (Type: string): A1 notation of the cell position.\\
- value: The value to set.\\
Returns:\\
- result (Type: object): The updated worksheet data or an error message.\\
\\
Name: update\_cell\_by\_formula(start\_position, end\_position, position\_list, result\_position, operator)\\
Description: Update the value of the target cell by applying formulas on some specified cells. Note: Either specify position\_list or start\_position and end\_position.\\
Parameters:\\
- start\_position (Type: string): The starting position of the range. Default: `B1'.\\
- end\_position (Type: string): The ending position of the range. Default: `D2'.\\
- position\_list (Type: array[string]): A list of cell positions in A1 notation.\\
- result\_position (Type: string): The position of the cell where the result of the formula will be stored in. Default: `G2'.\\
- operator (Type: string): The operator to be applied on selected cells. Choose one from [`SUM', `AVERAGE', `COUNT', `MAX', `MIN', `MINUS', `PRODUCT'].\\
Returns:\\
- result (Type: object): The updated worksheet data or an error message.\\
\\
Name: update\_range(start\_position, end\_position, values\_list)\\
Description: Update a range of the cells from a list\\
Parameters:\\
- start\_position (Type: string): A1 notation of the start cell.\\
- end\_position (Type: string): A1 notation of the end cell.\\
- values\_list (Type: array[array[Any]]): List of values to be inserted, which can be expressions\\
Returns:\\
- result (Type: object): The updated worksheet data or an error message.\\
\\
Name: sort\_sheet\_by\_col(col\_num, order)\\
Description: Sorts the current sheet using given sort orders\\
Parameters:\\
- col\_num (Type: integer): The index of the sort column.\\
- order (Type: string): The sort order. Possible values are `asc' or `des'.\\
Returns:\\
- result (Type: object): The updated worksheet data or an error message.\\
\\
Name: merge\_cells(start\_position, end\_position)\\
Description: Merge cells in sheet\\
Parameters:\\
- start\_position (Type: string): Starting cell position(top left) in A1 annotation.\\
- end\_position (Type: string): Ending cell position(bottom right) in A1 annotation.\\
Returns:\\
- result (Type: object): The updated worksheet data or an error message.\\
\bottomrule
\end{tabular}
\end{table*}

\begin{table*}[t]
\centering
\caption{Prompt details for Sheet (Part 3/4).}
\label{table:Prompt details for Sheet (Part 3/4).}
\vspace{2.8mm}
\begin{tabular}{p{0.96\linewidth}}
\toprule
Name: update\_note(position, content)\\
Description: Update a note in a certain cell\\
Parameters:\\
- position (Type: string): cell position in A1 annotation.\\
- content (Type: string): The text note to insert.\\
Returns:\\
- result (Type: string): The updated note or an error message.\\
\\
Name: get\_all\_values()\\
Description: Display all cell values in current sheet\\
Returns:\\
- result (Type: array[array[Any]]): Return all cell values or an error message.\\
\\
Name: get\_range\_values(start\_position, end\_position)\\
Description: Returns a list of cell data from a specified range.\\
Parameters:\\
- start\_position (Type: string): Starting cell position in A1 annotation.\\
- end\_position (Type: string): Ending cell position in A1 annotation.\\
Returns:\\
- result (Type: array[array[Any]]): List of cell data from the specified range or an error message.\\
\\
Name: get\_cell\_value(position)\\
Description: Get the value of a specific cell\\
Parameters:\\
- position (Type: string): Cell position in A1 annotation.\\
Returns:\\
- result : Cell value or an error message.\\
\\
Name: get\_value\_by\_formula(start\_position, end\_position, position\_list, operator)\\
Description: Calculate a value applying formulas on specified cells. Note: Either specify position\_list or start\_position and end\_position.\\
Parameters:\\
- start\_position (Type: string): The starting position of the range. Default: `B1'.\\
- end\_position (Type: string): The ending position of the range. Default: `D2'.\\
- position\_list (Type: array[string]): A list of cell positions in A1 notation.\\
- operator (Type: string): The operator to be applied on selected cells. Choose one from [`SUM', `AVERAGE', `COUNT', `MAX', `MIN', `MINUS', `PRODUCT'].\\
Returns:\\
- result (Type: string): Calculated result or an error message.\\
\\
Name: filter\_cells(query, in\_row, in\_column)\\
Description: Find all cells matching the query, return all cells' position.\\
Parameters:\\
- query (Type: [`string', `re.RegexObject']): A string to match or compiled regular expression.\\
- in\_row (Type: [`integer', `None']): Row number to scope the search. Default is all rows\\
- in\_column (Type: [`integer', `None']): Column number to scope the search. Default is all columns\\
Returns:\\
- result (Type: array[string]): List of cell addresses that match the query or an error message.\\
\\
Name: get\_note(position)\\
Description: Get the note at the certain cell, or return empty string if the cell does not have a note.\\
Parameters:\\
- position (Type: string): Cell position in A1 annotation.\\
Returns:\\
- result (Type: string): Note content or an error message.\\
\bottomrule
\end{tabular}
\end{table*}

\begin{table*}[t]
\centering
\caption{Prompt details for Sheet (Part 4/4).}
\label{table:Prompt details for Sheet (Part 4/4).}
\vspace{2.8mm}
\begin{tabular}{p{0.96\linewidth}}
\toprule
Name: finish()\\
Description: Return an answer and finish the task\\
Returns:\\
- result (Type: array[array[Any]]): Return all cell values or an error message.\\ \\
\textbf{Instruction:} \\
Now new trial starts. You should perform actions to accomplish the goal: Product Update: The table in ``Sheet1'' contains the product inventory information, and [[`Product', `Today Sold'], [`beef', `5'], [`pork', `2'], [`chicken', `8'], [`lamb', `12'], [`duck', `3'], [`fish', `23'], [`shrimp', `21'], [`salmon', `12'], [`apple', `100'], [`banana', `287'], [`orange', `234'], [`carrot', `12']] is today's sales data. Please update the product information in ``Sheet1'' in time and then sort by ``Quantity'' in descending order. Give me one action. \\
\bottomrule
\end{tabular}
\end{table*}

\end{document}